\documentclass[conference]{IEEEtran}
\IEEEoverridecommandlockouts
\usepackage{cite}
\usepackage{amsmath,amssymb,amsfonts}
\usepackage{graphicx}
\usepackage{textcomp}
\usepackage{xcolor}
\def\BibTeX{{\rm B\kern-.05em{\sc i\kern-.025em b}\kern-.08em
T\kern-.1667em\lower.7ex\hbox{E}\kern-.125emX}}

\usepackage{comment}
\usepackage{xspace}
\usepackage{caption}
\usepackage{epsfig}
\usepackage{tabularx}
\usepackage{multirow}
\usepackage[hidelinks]{hyperref}
\usepackage{booktabs}
\usepackage{enumitem}

\usepackage{algorithm}
\usepackage{algorithmicx}
\usepackage[noend]{algpseudocode}

\usepackage{ifthen}
\newboolean{supp}

\makeatletter
\DeclareRobustCommand\onedot{\futurelet\@let@token\@onedot}
\def\@onedot{\ifx\@let@token.\else.\null\fi\xspace}
\def\eg{\emph{e.g}\onedot} 
\def\ie{\emph{i.e}\onedot} 
 
\def\etc{\emph{etc}\onedot}

\makeatother

\newcommand{\ourmodel}{Scene-Actor Graph Neural Network\xspace}
\newcommand{\ourmodelshort}{\textsc{SA-GNN}\xspace}
\newcommand{\ourdataset}{\textsc{ATG4D}\xspace}
\newcommand{\nuscenes}{\textsc{nuScenes}\xspace}

\newcommand\given[1][]{\:#1\vert\:}

\newcommand{\suppref}{appendix}

\iftrue %
\newcommand{\cutsectionup}{\vspace*{-5pt}}
\newcommand{\cutsectiondown}{\vspace*{-3pt}}

\newcommand{\cutsubsectionup}{\vspace*{-3pt}}
\newcommand{\cutsubsectiondown}{\vspace*{-2pt}}

\newcommand{\cutsubsubsectiondown}{\vspace*{-0pt}}

\newcommand{\cutcaptionup}{\vspace*{-4pt}}
\newcommand{\cutcaptiondown}{\vspace*{-8pt}}

\newcommand{\cutequationup}{\vspace*{-2pt}}
\newcommand{\cutequationdown}{\vspace*{-0pt}}

\newcommand{\cuttabledown}{\vspace*{-8pt}}

\else %
\newcommand{\cutsectionup}{}
\newcommand{\cutsectiondown}{}

\newcommand{\cutsubsectionup}{}
\newcommand{\cutsubsectiondown}{}

\newcommand{\cutsubsubsectiondown}{}

\newcommand{\cutcaptionup}{}
\newcommand{\cutcaptiondown}{}

\newcommand{\cutequationup}{}
\newcommand{\cutequationdown}{}

\newcommand{\cuttabledown}{}

\fi

\begin{document}

\setboolean{supp}{false}

\bibliographystyle{IEEEtran}

\title{Safety-Oriented Pedestrian Motion and Scene Occupancy Forecasting}

\author{
\textbf{Katie Luo\thanks{* Denotes equal contribution (kzl6@cornell.edu, sergio@cs.toronto.edu)}$^{*,}$\thanks{$\dagger$ Work done while an AI resident at Uber ATG}$^{\dagger, 3}$, Sergio Casas$^{*,1,2}$, Renjie Liao$^{1,2}$} \\
\textbf{Xinchen Yan$^{1}$, Yuwen Xiong$^{1,2}$, Wenyuan Zeng$^{1,2}$, Raquel Urtasun$^{1,2}$} \\
Uber ATG$^1$, University of Toronto$^2$, Cornell University$^3$ \\
}

\maketitle

\begin{abstract}
In this paper, we address the important problem in self-driving of forecasting multi-pedestrian motion and their shared scene occupancy map, critical for safe navigation.
Our contributions are two-fold.
First, we advocate for predicting both the individual motions as well as the scene occupancy map in order to effectively deal with missing detections caused by postprocessing, \eg confidence thresholding and non-maximum suppression.
Second, we propose a \ourmodel (\ourmodelshort) which preserves the relative spatial information of pedestrians via 2D
convolution, and captures the interactions among pedestrians within the same scene, including those that have not been detected, via message passing.
On two large-scale real-world datasets, nuScenes and ATG4D, we showcase that our scene-occupancy predictions are more accurate and better calibrated than those from state-of-the-art motion forecasting methods, while also matching their performance in pedestrian motion forecasting metrics. 
\end{abstract}

\cutsectionup
\section{Introduction}
\cutsectiondown
Having access to an accurate predictive model of pedestrian behaviors with quantified uncertainties is critical to provide safe solutions to autonomous driving. 
Humans have a few priors about pedestrian behaviors. Pedestrians typically walk towards their desired destinations and generally follow sidewalks, avoid walls, etc., Furthermore,  pedestrians tend to interact with one another, either to avoid colliding with each other or walking together in groups~\cite{Helbing1995SocialFM,pellegrini2009you}.
While these are good priors to incorporate in motion forecasting models, the erratic nature of pedestrian motion in urban settings makes it challenging to model, and using a simply parameterized output such as trajectories with Gaussian uncertainty often yields to unrealistic behaviors \cite{alahi2016social,jain2019discrete}.

A better alternative is to exploit non-parametric representations \cite{jain2019discrete,sadat2020perceive} that can naturally capture multi-modal behaviors. For instance, a pedestrian reaching a crosswalk may decide to cross, wait at the intersection, or continue walking along the sidewalk. Urban environments also influence the timing of the pedestrian motion. For example, a traffic light is (usually) obeyed by pedestrians, and the changes in the state of the light will dictate the moment they start moving. However, pedestrians might also decide to perform illegal actions such as jaywalking, and these are especially important to estimate for safety reasons.

Joint perception and prediction approaches \cite{luo2018fast,casas2018intentnet,zeng2019end,liang2020pnpnet} predict object detections and motion
forecasts from onboard sensor data, effectively dealing with sensor noise and occlusions while sharing computation between tasks.
This stands in contrast to most works in pedestrian motion forecasting \cite{alahi2016social,sadeghian2019sophie, kosaraju2019social, mohamed2020social}, which assume perfect perception. 
However, these joint perception and prediction efforts focus on vehicles, for which detection is stronger than that of pedestrians since (i) vehicles are larger and thus get much more evidence from LiDAR returns and (ii) driving datasets are dominated by vehicles.
Furthermore, object detection involves a discrete decision to produce a final set of bounding boxes, which is typically achieved through confidence thresholding and non-maximum suppression (NMS). These postprocessing operations can discard detections of true pedestrians because of their low confidence, such as a partially occluded pedestrian among parked vehicles that is about to jaywalk in the road. %
This postprocessing poses a threat to safe navigation as it limits perception recall. %

To tackle missing detections, our framework combines the motion predictions for all the pedestrians that have been detected with a scene-level occupancy map predicted from dense features in an instance-free manner (i.e., without any postprocessing). This way the object detector can be operated at a high precision point, without the safety concern of missing a pedestrian as this will be recovered by the scene-level occupancy. 
More precisely, we design a perception and prediction model that is trainable end-to-end and features a \ourmodel (\ourmodelshort) that models interactions not only between detected actors but also with the occupancy of the scene as a whole.

We showcase the power of this approach 
on \nuscenes~\cite{caesar2019nuscenes} and \ourdataset~\cite{yang2018pixor}. 
Our work achieves significant improvements over state-of-the-art motion forecasting methods in the task of occupancy forecasting while achieving on par results on instance-based motion forecasting metrics.
With our work, we hope to motivate more research on scene-occupancy prediction, showing that such a failsafe system is critical for safety applications to achieve high recall, as the aggregation of independent predictions per agent are shown to not be sufficient in our experiments.

\cutsectionup
\section{Related work}
\cutsectiondown

While there exist many works for perception and prediction of vehicles \cite{luo2018fast,casas2018intentnet,casas2019spatially,casas2020implicit}, only a few explored the topic of pedestrian motion forecasting in a real-world setting where no perfect detections and past tracks are assumed \cite{liang2020pnpnet, sadat2020perceive}, despite the safety-critical nature of correctly perceiving them and accurately predicting their motion. We begin by highlighting a few existing works on LiDAR-based perception, and then explore previous pedestrian motion forecasting methods.

\textbf{Object Detection from LiDAR:}
The availability of LiDAR sensory data has advanced object detection in robotic applications and autonomous driving systems.
Several existing works \cite{song2016deep,li20173d,engelcke2017vote3deep,luo2018fast,zhou2018voxelnet,zhou2019end} grouped the LiDAR point clouds into voxelized grids and adapted 3D convolutional neural networks for predicting bounding box proposals. 
These approaches achieve high performance but are inefficient as they require heavy computation and memory cost to process the sparse LiDAR data.
PointNet~\cite{qi2017pointnet,qi2017pointnet++,qi2018frustum,shi2019pointrcnn} and Graph-based Networks~\cite{qi20173d,wang2018deep} have been explored to learn discriminative representations directly from the point clouds (without voxelization). Albeit conceptually attractive, point-based approaches have achieved limited performance.
Efforts have also been made in projecting sparse LiDAR point clouds to bird's eye view (BEV) so as to learn 3D bounding box proposal using traditional 2D convolutional neural networks~\cite{chen2017multi,ku2018joint,yang2018pixor,liang2018deep,yang2018hdnet,yang2019std,liang2019multi}.
These methods have achieved a great tradeoff between predictive power and real-time performance, and this is the framework we start off from.
Compared to the existing work, however, our proposed {\ourmodelshort} also considers the instance-free, scene-level occupancy (apart from the actor-level perception), thus being robust to detection failures.

\textbf{Multi-agent Motion Prediction with Deep Neural Networks:}
Motion prediction in the multi-agent setting has been studied in vision and learning community for decades~\cite{Helbing1995SocialFM,ridel2018review,Rudenko2019HumanMT}.
Recent works~\cite{becker2018,alahi2016social,gupta2018social,xu2018encoding,zhu2019starnet} formulated the task as sequence prediction problem and explored the long short-term memory (LSTM) \cite{hochreiter1997long} networks to model the temporal dependencies.
Others \cite{li2017situation,schlichtkrull2018modeling,kipf2018neural,zhang2019stochastic,sadeghian2019sophie,kosaraju2019social} applied (Graph Neural Network) GNNs~\cite{scarselli2008graph} to capture interactions among multiple agents.
Despite the large body of works, only a handful of them explored the topic in the context of self-driving~\cite{luo2018fast,casas2018intentnet,casas2019spatially,jain2019discrete,chai2019multipath,tang2019multiple,casas2020importance,casas2020implicit}.
In particular, \cite{luo2018fast} introduced a model that unifies object detection and short-term trajectory forecasting of vehicles from LiDAR sensory data.
Building upon this framework, \cite{casas2019spatially} proposed to model multi-agent interactions using spatially-aware graph neural networks.
We exploit this framework to perceive and forecast pedestrian behavior, but leverage non-parametric spatial distributions to predict the individual pedestrian's future motion \cite{jain2019discrete} as well as the scene occupancy \cite{sadat2020perceive}.
Moreover, our {\ourmodelshort} captures the interaction among detected pedestrians and the scene-level occupancy map which significantly improves the safety and robustness.
We note that our problem setup is very different from most previous works in pedestrian or vehicle motion forecasting \cite{alahi2016social,sadeghian2019sophie, kosaraju2019social, mohamed2020social, jain2019discrete,tang2019multiple}, which assume perfect perception, i.e. that the ground-truth past trajectories of each actor are given. 
This is unsafe for self-driving since it neglects potential perception failures such as false positive,  false negative detections as well as tracking fragmentations caused by occlusion and sensor noise.
In contrast, we directly consider the sensor data and HD maps as input and deal with perception failures.

\begin{figure*}
    \begin{center}
    \includegraphics[width=\linewidth]{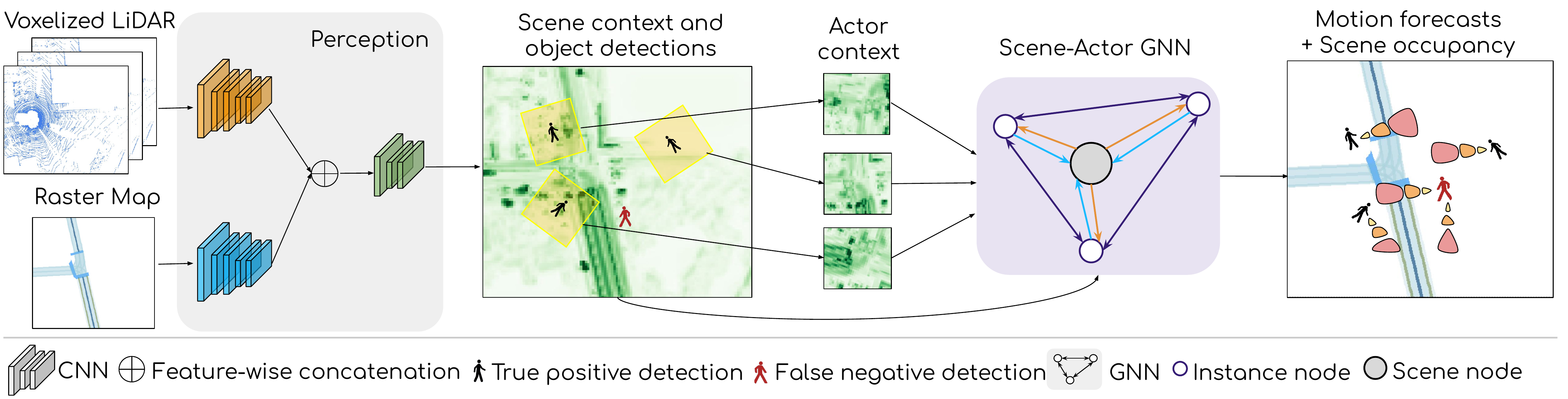}
    \end{center}
	\cutcaptionup
       \caption{Our model perceives pedestrians (Section \ref{subsec:perception}) and predicts their current and future occupancy (Section \ref{subsec:prediction}) from LiDAR and HD maps. Notably, the proposed model can recover from false negative detections and still predict their occupancy over time.}
    \cutcaptiondown
    \cutcaptiondown
    \label{fig:overall}
\end{figure*}

\cutsectionup
\section{Forecasting Pedestrian Motion and Scene Occupancy}
\cutsectiondown

Detecting individual pedestrians and forecasting their future motion gives the downstream task of motion planning an interpretable representation of free-space over time, where it can verify that product and regulation requirements such as "slowing down when a pedestrian is running towards the road" are met. 
However, the object detector may miss actual pedestrians (i.e. produce false negatives) due to the confidence thresholding or non-maximum suppression
(NMS), which would pose severe safety concerns especially when this is the only input information to an SDV's motion planner.
For this reason, we also forecast the occupancy of the scene as a whole \cite{milan2016online, hoermann2018dynamic, mcgill2019probabilistic, sadat2020perceive}. 
In this representation, we lose the notion of an instance (or actor), but we can achieve full recall and be safe. 
Moreover, we unify both tasks in a single end-to-end model. This helps by (i) reusing computation from the per-instance reasoning to predict the scene, and (ii) by improving the individual motion forecasts in the case where a detected pedestrian is interacting with another pedestrian that has not been recalled by the object detector.
We interchangeably use \emph{pedestrian}, \emph{actor} or \emph{instance}, as pedestrian is the only actor category we consider in this paper.

The overall architecture of our approach is illustrated in Fig. \ref{fig:overall}. 
We first extract global features from voxelized LiDAR point clouds and a raster map using CNNs and perform object detection.
We then extract a local context for each pedestrian and build an interaction graph where each detected actor is a node and the context feature of the scene is the supernode in the graph.
Detected pedestrians are connected based on spatial proximity, and all actors are connected to the scene node.
To reason efficiently and effectively on this graph, we employ GNNs, and design a message passing process using 2D convolution in order to preserve spatial information for modeling the interactions among pedestrians, unlike typical GNNs that work with feature vectors. 
After message passing, each node (including the supernode) makes a prediction, thus obtaining instance motion forecasts and the shared scene occupancy.

In the remainder of this section, we first describe our perception module including the input representation, then explain our unified \ourmodelshort as well as the output parameterization for motion and occupancy, and finally our end-to-end learning framework.

\cutsubsectionup
\subsection{Pedestrian Perception from LiDAR and HD Maps}
\label{subsec:perception}
\cutsubsubsectiondown
We now discuss the perception module, which exploits LiDAR point clouds and HD maps as input, detects the pedestrians, and provides the scene and actor context features to the prediction module.

\textbf{LIDAR:}
In order to exploit motion history to estimate future behavior, we follow \cite{luo2018fast} and leverage multiple LiDAR sweeps ($N_\text{sweep}$) by projecting the past sweeps to the the current coordinate frame (i.e. correcting for ego-motion).
Then, we voxelize the 3D point clouds in Bird's Eye View (BEV) and concatenate the depth and time dimensions of our tensor in order to exploit 2D convolutions, which have been shown to be very effective \cite{casas2018intentnet,zeng2019end}. 
This results in a 3D tensor of dimensions $(H, W, D \cdot N_\text{sweep})$, with $H, W, D$ being the height, width, and depth dimensions in BEV.

\textbf{HD Map:}
Following \cite{casas2018intentnet}, we rasterize the map to include binary maps that encode different semantics in separate channels. 
These contain information including lanes, roads, intersections, crossings, traffic lights, \etc for a total of 17 binary channels. 
By encoding the information into separate channels, we are able to ease the learning and avoid predefining orderings in the raster.

\textbf{Backbone Network and Object Detection Header:}

Our backbone network is a two-stream convolutional neural network (CNN) inspired by \cite{yang2018pixor}, with one stream processing LiDAR point clouds and the other processing HD maps.
Features from both streams are then concatenated along the channel dimension and fused by another CNN header. 
We refer to the fused feature as \emph{scene context feature}, denoted as $\mathcal{C}$, which is used by both the detection and the prediction modules. 
For detection, these features are fed to two separate CNN headers that output a confidence score, and offsets for centroid and heading for each anchor pixel. 
Finally, these outputs are reduced to the final set of candidate bounding box detections $\mathcal{B}$ by applying confidence thresholding and non-maximum suppression (NMS).

\cutsubsectionup
\subsection{Scene-Actor Graph Neural Network}
\label{subsec:prediction}
\cutsubsectiondown

We begin with the precise definition of scene occupancy and actor motion representations and then introduce the prediction module which takes in the scene context feature $\mathcal{C}$ and detections $\mathcal{B}$ from the perception module and outputs the future scene occupancy and motion of each individual actor.

\textbf{Occupancy and Motion Representations:}
We introduce two types of coordinate systems: a scene frame shared by all actors and an actor-centric frame for each actor.
For the scene occupancy, we use a scene frame that is relative to the pose of the self-driving vehicle and discretize a rectangular region into an $H_{S} \times W_{S}$ spatial grid, where $H_{S}$ and $W_{S}$ are the number of grid cells in the scene y-axis and x-axis respectively. 
At any future timestep $t$, we denote $S^{t}$ as a collection of bernoulli random variables, where  $S^{t}_{x,y} = 1$ indicates  that   position $(x,y)$ is occupied by some actor.
For the $i$-th detected actor, we discretize the space around it in an actor-centric frame defined by the detection bounding box $\mathcal{B}_i$ centroid and heading.
We subdivide a rectangular region of interest (RoI) around each actor into $N$ grid cells, where $N = h_{I} \times w_{I}$, $h_{I}$ and $w_{I}$ are the height and width of the RoI, respectively. 
The motion of the $i$-th actor at time $t$ is represented as $I_{i}^{t}$ which is a categorical random variable that can take on $N$ possible values indicating the event that this actor moves to a particular location.
Throughout the paper, we consider the future time horizon up to $T$ timesteps.

\textbf{Probabilistic Formulation:}
We are interested, for the scene-level occupancy,  in computing the probability of a given location being occupied by any pedestrian at a future time step, \ie, $P(\{S^t \vert t = 0, \cdots, T \} \vert \mathcal{C})$.
For the actor motion, we'd like to compute the probability of a given actor being at a particular location in the future, \ie, $P(\{I_{i}^{t} \vert i = 1, \cdots, N, t = 0, \cdots, T \} \vert \mathcal{C})$.
To make the model tractable, we adopt the following assumptions.
First, we assume that conditioned on the context $\mathcal{C}$, the scene occupancy map can be predicted independently among different time steps, \ie, $P( \{S^t \vert t = 0, \cdots, T \} \vert \mathcal{C} ) = \prod_{t} P( S^t \vert \mathcal{C} )$.
Second, we assume that if conditioned on the context $\mathcal{C}$, whether or not a pedestrian will be at any location is conditionally independent of the other pedestrians, \ie, $P(\{I_{i}^{t} \vert i = 1, \cdots, N, t = 0, \cdots, T \} \vert \mathcal{C}) = \prod_{i}\prod_{t} P( I_{i}^{t} \vert \mathcal{C} )$.
Note that these conditional independence assumptions are reasonable since the context tensor summarizes the history of all pedestrians and their surroundings.
Also, \cite{jain2019discrete,casas2019spatially,chai2019multipath,casas2020importance} showed that there was barely any gain from modeling temporal dependencies and it comes at the expense of runtime. 
To capture the interaction among multiple pedestrians, we construct our model using graph neural networks (GNNs), which have been shown to perform well on probabilistic inference tasks \cite{yoon2018inference}. 
In the following, we will explain how we build the graph and then introduce the core part of our {\ourmodelshort}, \ie, message passing process.

\begin{algorithm*}[h]
    \caption{Message passing in \ourmodelshort}
    \cutcaptiondown
    \label{alg:msg_passing}
    \begin{flushleft}
    \textbf{Input:}
    Initial state for all actors $\begin{Bmatrix}h^{0}_{I_i} : i \in 1 \dots N\end{Bmatrix}$ and scene $h^{0}_{S}$. Interaction graph $G = (V, E)$. 

    \end{flushleft}
    \begin{algorithmic}[1]
    \For {$ k \in 1 \dots K $} \Comment{\small {(sequential)}}
        \For {$ (i, j) \in E $} \Comment{\small {(in parallel)}}
            \State $m_{I_j \rightarrow I_i}^{k} = F_{\text{actor}} \big( F_{\text{emb}} \big( h_{I_i}^{k-1}) \parallel \mathcal{T}_{I_j \rightarrow I_i} \big( F_{\text{emb}} ( h_{I_j}^{k-1} \big) \big) \big)$ \Comment{\small {Compute \textit{actor-to-actor} messages}}
        \EndFor
        \For {$ i \in V $}  \Comment{\small {(in parallel)}}
            \State $h^{k}_{I_i} = U_{\text{actor}} \big( F_{\text{deconv}} \big(\texttt{max} \big( \big\{ m_{I_j \rightarrow I_i}^{k} : j \in \mathbf{N} (i) \big\}\big)\big) \parallel h^{k-1}_{I_i} \big)$   \Comment{{\small Update actor states}}
            \State $m_{I_i \rightarrow S}^{k} = \mathcal{T}_{I_i \rightarrow S} \big( F_{\text{instance}}\big(h_{I_i}^{k} \big)\big)$ \Comment{{\small Compute \textit{actor-to-scene} messages}}
        \EndFor
        \State $h^{k}_{S} = U_{\text{scene}} \big(\texttt{max} \big( \big\{ m_{I_i \rightarrow S}^{k} : \forall \text{ actor } i \big\}\big) \parallel h^{k-1}_{S} \big)$  \Comment{{\small Update scene node state}}
        \For {$ i \in V $}  \Comment{\small {(in parallel)}}
        \State $ m_{S \rightarrow I_i}^{k} = F_{\text{scene}} \big( \texttt{RoIAlign}_{i} \big( h_{S}^{k-1} \big) \parallel h_{I_i}^{k} \big)$  \Comment{{\small Compute \textit{scene-to-actor} messages}}
        \State $h^{k}_{I_i} = U_{\text{actor}} \big( m_{S \rightarrow I_i}^{k} \parallel h^{k}_{I_i} \big)$  \Comment{{\small Update actor states with scene information}}
        \EndFor
    \EndFor
    \State\Return $\begin{Bmatrix}h^{K}_{I_1}, h^{K}_{I_2}, \cdots, h^{K}_{I_N}\end{Bmatrix}$, $h^{K}_{S}$   \Comment{{\small Return updated node states}}
    \end{algorithmic}
\end{algorithm*}

\textbf{Interaction Graph:}
We  build an \textit{interaction graph} to effectively capture  interactions between pedestrians.
Specifically, we treat each actor as a node and connect two nodes  if (1) the distance between pedestrians is less than $32$ meters and (2) the candidate pedestrian is among the top-$K$ closest neighbours. We choose $K = 6$ in our experiments due to GPU memory constraints.
Additionally, we regard the scene level occupancy map as a super node which connects to all actors in the scene.
With this graph construction, the scene-level occupancy information can be easily propagated to the actor nodes.
This way even if the perception module fails to detect some actors, the occupancy information can still enable detected actors to be aware of such false negatives. 
Most importantly, the downstream task of motion planning can reason about future occupancy even for undetected actors.

\textbf{Message Passing:}
Given the interaction graph, the scene node state $h_{S}^{0}$ is initialized using the context feature $\mathcal{C}$.
Similarly, we initialize the $i$-th actor node's state $h_{I_i}^{0}$ via the RoI feature extracted from $\mathcal{C}$ via \texttt{RoIAlign}~\cite{he2017mask}.
Here, the superscript denotes the message passing step.
Since within the RoI of a pedestrian there might be multiple other pedestrians, it is desirable to make the 2D convolutions translation variant, for which we use CoordConv \cite{liu2018intriguing} in the actor-centric frame. 
For ease of notation, we assume that the coordinate layers are included in $h_{I_i}^{0}$.
To preserve the spatial information in graph representations and predict socially consistent future occupancy, we integrate the 2D spatial convolution into the design of {\ourmodelshort}.
Algorithm~\ref{alg:msg_passing} provides pseudo-code for our finite step message passing process, which we describe next.

We first compute the \emph{actor-to-actor} messages, where $m_{I_j \rightarrow I_i}^{k}$ denotes the message from actor $j$ to $i$.
$\mathcal{T}_{I_j \rightarrow I_i}$ is the bilinear resampling function defined by the affine transformation of actor $j$ into actor $i$'s coordinate frame.
$F_{\text{emb}}$ is the node embedding network, implemented as a shallow fully convolutional network (FCN) that keeps the spatial dimensions from the RoI.
$F_{\text{actor}}$ is the message network, which takes as input the concatenation of the neighboring node states along the feature dimension (denoted by $\parallel$), and applies a shallow CNN with large strides to obtain a feature vector. 
Then we aggregate the messages from neighboring actors ($\mathbf{N} (i)$ is the set of neighboring actors of actor $i$) via the element-wise \texttt{max} operator, then we recover the spatial dimensions via transposed convolutions $F_{\text{deconv}}$, and finally update the state via $U_{\text{actor}}$ which is instantiated as a ConvGRU.

Second, we compute \emph{actor-to-scene} messages by first processing the actor's (spatial) state with another shallow FCN ($F_{\text{instance}}$) and then warping this feature map into the coordinate frame of the scene using the affine transformation $\mathcal{T}_{I_i \rightarrow S}$ and pasting it into a canvas (\ie, initialized as a zero tensor that has the same spatial size as the scene) accumulatively.
We aggregate messages from all actors via \texttt{max} and update the scene state via $U_{\text{scene}}$ which is again a ConvGRU.

Last, we compute \emph{scene-to-actor} messages by first applying \texttt{RoIAlign} to extract features from scene state and concatenating it with the recently updated node state.
Then we pass the concatenated feature to the scene message function $F_{\text{scene}}$ (implemented as another shallow FCN) and update the actor state again.

\textbf{Output:}
After running $K$ steps of the above message passing, we predict the actor motion as $p(I_{i}^{t} \given \mathcal{C}) = O_{\text{actor}} ( h_{I_i}^{K} )$ and then aggregate the conditionally independent predictions into the scene as $P_{\text{agg}}( S^t \vert \mathcal{C} ) = 1 - \prod_i (1 - p'(I_{i}^{t}))$, where $p'$ denote the bilinearly interpolated instance probabilities into the scene frame.
Finally, we predict the scene occupancy map as $P( S^t \vert \mathcal{C} ) = O_{\text{scene}} ( h^{K}_{S}, P_{\text{agg}}( S^t \vert \mathcal{C} ))$.
Both $O_{\text{actor}}$ and $O_{\text{scene}}$ are implemented as shallow FCNs.

\cutsubsectionup
\subsection{End-to-end Learning}
\cutsubsectiondown
We train our whole system in an end-to-end fashion via back-propagation. 
Our final loss is a linear combination of the two groups of losses, \ie, instance and scene losses:
\begin{equation}
{\small
    \mathcal{L}_{total} = \underbrace{\lambda_1 \mathcal{L}_{\text{class}} + \lambda_2 \mathcal{L}_{\text{reg}} + \lambda_3 \mathcal{L}_{\text{pred}}}_{\text{Instance Loss}} + \lambda_4 \underbrace{ \sum\nolimits_{t=1}^{T} \left( - \log P (S^{t} \given \mathcal{C}) \right)}_{\text{Scene Loss}},
}
\end{equation}
where the instance losses further decompose into classification loss of detection $\mathcal{L}_{\text{class}}$, regression loss of detection offsets $\mathcal{L}_{\text{reg}}$ and negative log likelihood of future motion $\mathcal{L}_{\text{pred}}$.
The coefficients $\{\lambda_i\}$ are determined via cross validation. 
Details of the these losses and hyperparameters can be found in the \suppref.

\cutsectionup
\section{Experiments}
\cutsectiondown
\label{sec:evaluation}

In this section, we empirically verify the effectiveness of our method.
In the following, we first introduce the datasets with their associated task details, then the metrics, and finally results including a comparison against state-of-the-art methods as well as extensive ablation studies.

\ifthenelse{\boolean{supp}}{
    See our supplementary materials \footnote{Supplementary materials: \hyperlink{supplementary materials}{https://uber.box.com/v/icra21-pedpnp-supp}} for implementation details and additional results.
}{
    See our appendix for implementation details and additional results.
}

\cutsubsectionup
\subsection{Datasets}
\cutsubsectiondown

We benchmark our method on two large-scale real-world datasets, our \ourdataset \cite{yang2018pixor} and \nuscenes \cite{caesar2019nuscenes}. We defer the experiments on \nuscenes to the \suppref. %

\ourdataset consists of more than one million frames over several cities in North America with a 64-beam, roof-mounted LiDAR sensor.
There are 6,500 snippets in total, each 25 seconds long, accounting for a total of about 500,000 annotated pedestrian trajectories. 
We have access to high definition maps capturing the geometry and topology of each road network.
Our pedestrian labels are precise 3D tracks over time with a maximum distance from the self-driving vehicle of 100 meters, enabling long term motion forecasting.
In particular, in this dataset we predict the future 7 seconds.
Our evaluations are on a held out validation set with over 34,000 trajectories.

\cutsubsectionup
\subsection{Metrics}
\cutsubsectiondown

We evaluate our results on two sets of metrics: (i) scene occupancy metrics and (ii) actor motion forecasting metrics.

\textbf{Scene Occupancy Metrics:}
We use the \emph{Precision-Recall (PR) curve} to evaluate the quality of the forecasted scene occupancies over time, given that each pixel is a Bernoulli random variable. We also use its mean average precision (mAP) to summarize the curve with a single scalar.
The \emph{PR curve} gives as an idea of how well the method ranks occupancy, but it does not reflect if the probabilities are well calibrated. Thus, we report a \emph{reliability diagram} \cite{guo2017calibration} that shows how well confidence correlates with accuracy. This is very important for safety-critical robotic applications like self-driving where a decision must be made taking into account the estimated probability of different events. From this plot, we can extract the (macro) average calibration error (ACE) and maximum calibration error (MCE).
To compare with previous motion forecasting approaches that operate only at the instance level, we aggregate their results to form the scene occupancy prediction as described in the \suppref.

\newcolumntype{s}{>{\centering\arraybackslash}X}

\begin{table*}[t]
	\centering
    \begin{tabularx}{\textwidth}{l|ssss|ssss}
        \toprule
        Model &  \multicolumn{4}{c|}{\textbf{Scene Occupancy}} &  \multicolumn{4}{c}{\textbf{Actor Motion Forecasting}} \\
        &  Avg. mAP$\uparrow$  &  Final mAP$\uparrow$ &  ACE$\downarrow$ & MCE$\downarrow$ &  NLL$\downarrow$ &  Exp. RMSE$\downarrow$ &  Argmax RMSE$\downarrow$ &  MinRMSE$\downarrow$ @25 \\
        \midrule
        STGAT \cite{huang2019stgat}                     &  17.63    &  5.34      &  7.77     &  22.11     & 5.28     &  3.75     &  2.26     &  1.15      \\
        SocialSTGCNN \cite{mohamed2020social}           &  32.41    &  12.71     &  9.85      &  25.40     & 4.78     &  2.65     &  1.74     &  0.94      \\
        SpAGNN \cite{casas2019spatially}                &  33.51    &  12.59     &  7.99      &  15.99     & 4.67     &  2.59      &  1.73     &  0.88      \\
        MultiPath \cite{chai2019multipath}              &  35.09    &  16.60     &  11.06     &  18.70     & 4.18     &  2.30     &  1.71     &  0.86      \\
        DRF-Net \cite{jain2019discrete}                 &  32.00    &  15.53     &  5.74      &  10.40     & 4.11     &  \textbf{2.26}     &  1.72     &  0.75      \\
        \ourmodelshort (Ours)                                 &  \textbf{37.69}      &    \textbf{19.28}   &    \textbf{1.21}  &  \textbf{2.55}    &    \textbf{4.02} &             2.33 &           \textbf{1.70}       &  \textbf{0.73}     \\
        \bottomrule
    \end{tabularx}
    \caption{\textbf{[\ourdataset{}] Scene occupancy and motion forecasting evaluation}.}
    \cuttabledown
    \label{table:atg4d}
\end{table*}

\textbf{Motion Forecasting Metrics:}
We use a comprehensive set of metrics to measure the different aspects of the motion distributions at different future horizons.
We measure the negative log-likelihood (\emph{NLL}) at the ground-truth location, which favours coverage over precision \cite{rhinehart2018r2p2}.
To measure how well our method ranks, we use
\emph{MinRMSE@M}, which computes the minimum $\ell_2$ distance from any of the top-$M$ most likely grid cells' center to the ground-truth. 
\emph{ArgmaxRMSE} is the case where $M=1$.
Finally, to measure the precision of the overall distribution, we propose \emph{Expected RMSE}, which is a weighted sum of the $\ell_2$ distances between the ground-truth and each cell's center, weighted by the probability the model predicts for the cell.
Since we perform joint perception and prediction, these metrics can only be evaluated on true positive detections. 
For a fair comparison, we follow \cite{casas2019spatially} and evaluate all models at a common recall point (70\% recall for \ourdataset).
To get a full picture on how our model compares to state-of-the-art motion forecasting approaches, we also consider methods that have different output parameterizations.
For methods~\cite{mohamed2020social,chai2019multipath,casas2019spatially} that treat the motion as the continuous trajectory and predict the probability density function (pdf), we approximate the integral of the pdf over the grid cells to obtain the probability mass function for evaluation.
For methods~\cite{huang2019stgat} that output trajectory samples without predicting the pdf, we first estimate the pdf using kernel density estimation (KDE), and then integrate that over the grid cells. 
More details are explained in the \suppref.

\cutsubsectionup
\subsection{Results}
\cutsubsectiondown

\begin{figure}[t]
    \centering
    \begin{tabular} {@{\hspace{.1em}}c@{\hspace{.5em}}c}
        {\includegraphics[width=0.53\columnwidth]{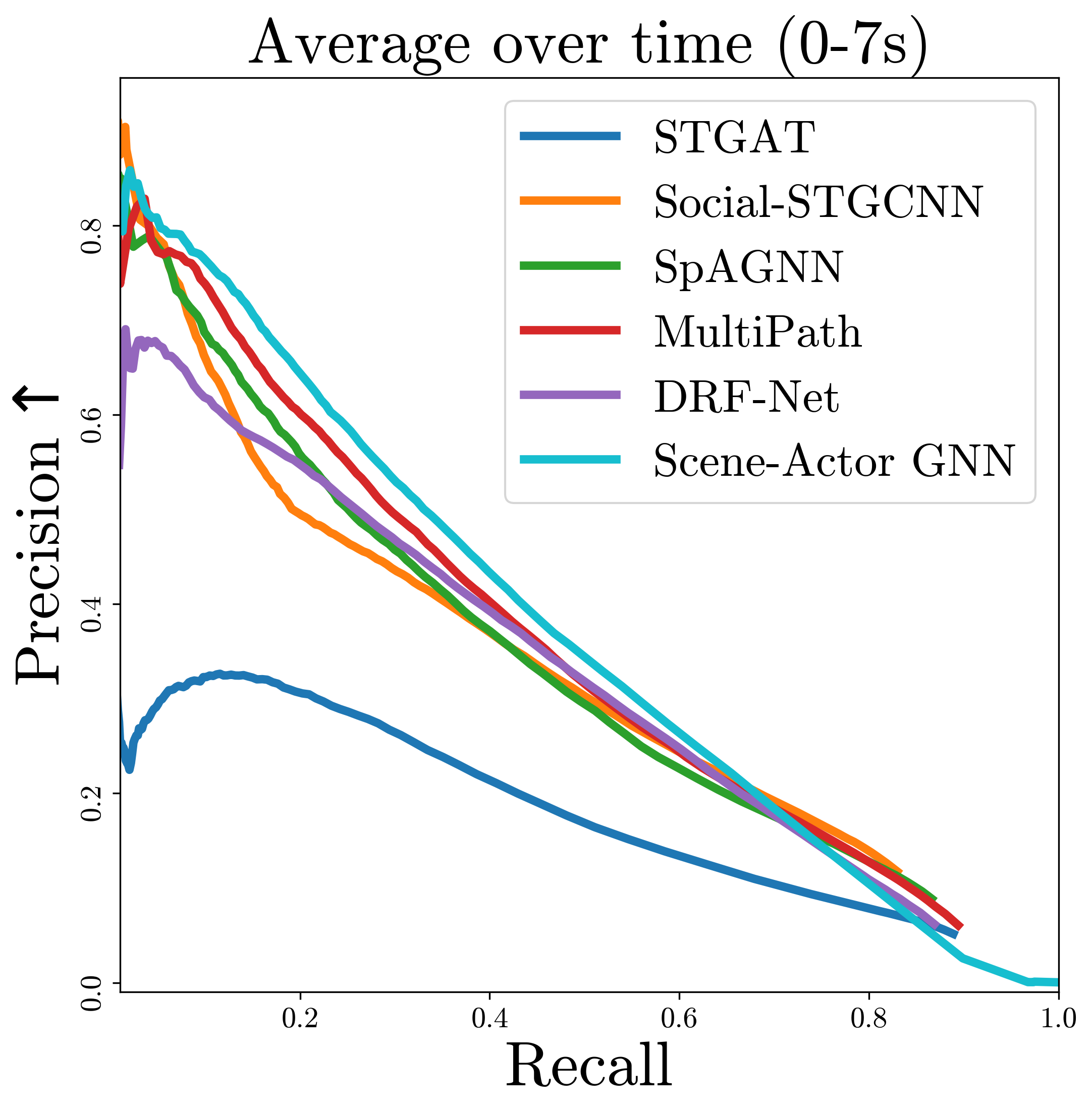}} &
        {\includegraphics[width=0.45\columnwidth]{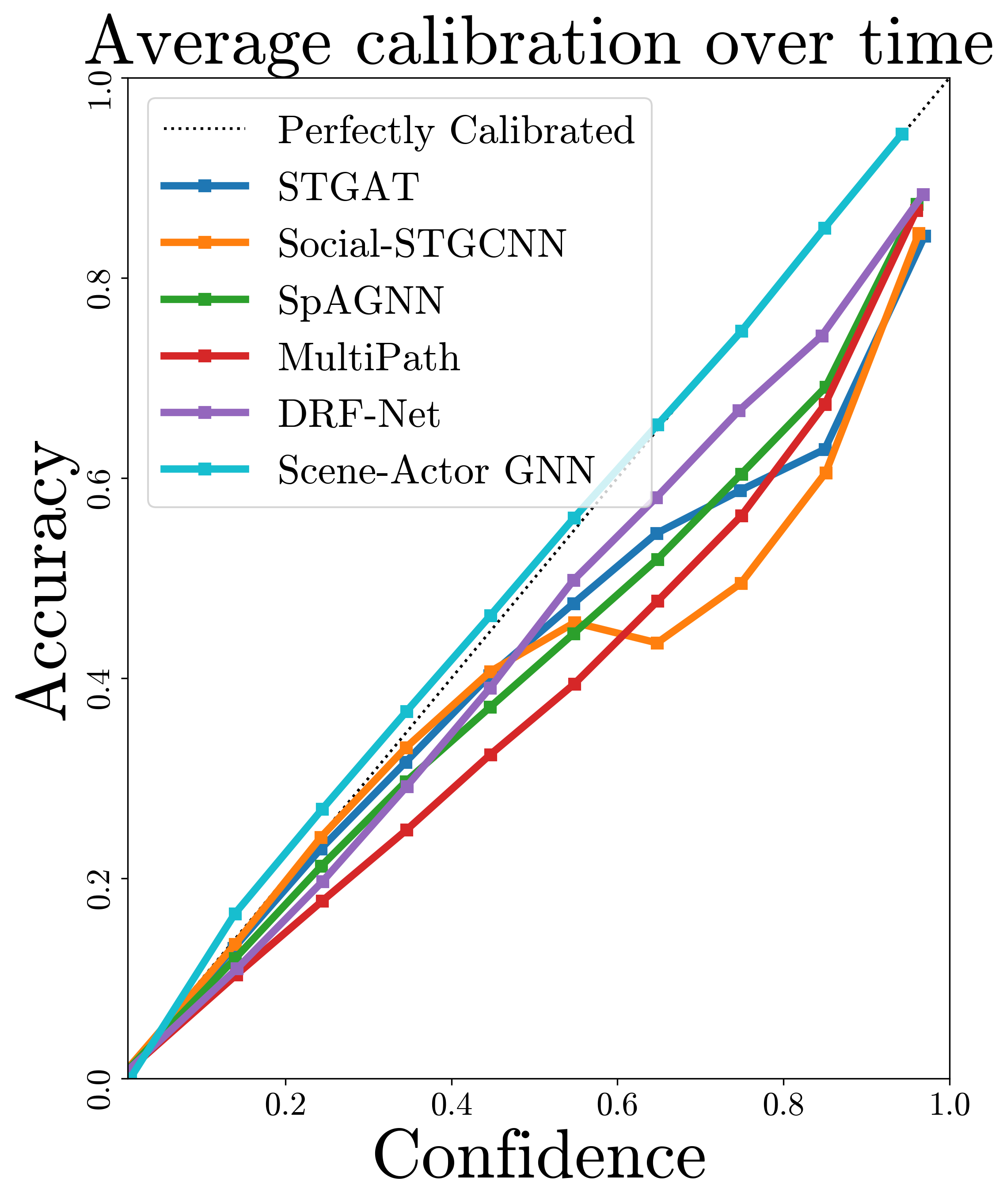}} \vspace{.5em}
    \end{tabular}
    \cutcaptionup
    \caption{\textbf{[\ourdataset{}] Scene occupancy evaluation}. PR curves and reliability diagram. Note that the baselines do not achieve 100\% recall due to detection post-processing.}
    \cutcaptiondown
    \label{fig:scene_atg4d}
\end{figure}

We benchmark our \ourmodelshort against the following set of state-of-the-art models in instance-level motion forecasting and scene-level occupancy metrics: 
SpAGNN \cite{casas2019spatially},
SocialSTGCNN \cite{mohamed2020social},
STGAT \cite{huang2019stgat},
MultiPath \cite{chai2019multipath},
and DRF-Net \cite{jain2019discrete}.

\textbf{Scene-level Occupancy Comparison:}
We compare our model performance on the scene-occupancy prediction task, critical for safe motion planning.
As shown in Fig.~\ref{fig:scene_atg4d} and Table~\ref{table:atg4d} our model \ourmodelshort outperforms all the baselines with stronger precision-recall curves and more reliable calibration.
In particular, we highlight that our model is able to achieve (i) a longer tail of recall thanks to the absence of detection thresholding on the scene-level occupancy, (ii) higher precision at low recall regimes, and (iii) a solid reliability diagram (near perfect in \ourdataset) without any postprocessing of the output probabilities.
We emphasize the importance of scene occupancy performance towards safe self-driving, where a probabilistic representation of free space over time that can capture low-probability events is a must. 

\newcolumntype{s}{>{\centering\arraybackslash}X}

\begin{table*}[t]
	\centering
    \begin{tabularx}{\textwidth}{l|ssss|ssss}
        \toprule
        
        Model &  \multicolumn{4}{c|}{\textbf{Scene Occupancy} } &  \multicolumn{4}{c}{\textbf{Actor Motion Forecasting}} \\
        {} &  Avg. mAP$\uparrow$  &  Final mAP$\uparrow$ &  ACE$\downarrow$ & MCE$\downarrow$ &  NLL$\downarrow$ &  Exp. RMSE$\downarrow$ &  Argmax RMSE$\downarrow$ &  MinRMSE$\downarrow$ @25 \\
        \midrule
        FCN                         &  32.84      &    15.71   &    4.76 &  9.15   &    4.55 &             3.40 &           2.42       &  0.92     \\
        FCN+CoordConv               &  33.47      &    16.31   &    4.44 &  6.02   &    4.24 &             2.44 &           1.83       &  0.78    \\
        \ourmodelshort w/o scene                &  34.02      &    16.78   &    2.54  &  5.92   &    4.06 &             \textbf{2.32} &           1.72       &  0.74     \\
        \ourmodelshort w/o s2a               &  \textbf{38.42}      &    \textbf{20.78}   &    1.34  &  3.05   &    4.27 &             2.46 &           1.84       &  0.78     \\
        \ourmodelshort (Ours)             &  37.69      &    19.28   &    \textbf{1.21}  &  \textbf{2.55}    &    \textbf{4.02} &             2.33 &           \textbf{1.70}       &  \textbf{0.73}     \\
        \bottomrule
    \end{tabularx}
    \caption{\textbf{[\ourdataset{}] Ablation study} for scene occupancy and instance-level motion forecasting.
}
    \cuttabledown
    \cuttabledown
    \label{table:atg4d_ablation}
\end{table*}

\textbf{Instance-level Motion Forecasting Comparison:}
Table \ref{table:atg4d} also shows a comparison with state-of-the-art in instance-level metrics. All numbers are for the final time step predictions (7s in \ourdataset). 
Our method performs equally or better than previous methods across all metrics, even though this is not the main goal of our work.

\textbf{Ablation Study:} 
Table~\ref{table:atg4d_ablation} shows a breakdown of the contributions of our \ourmodelshort.
We start off with a fully convolutional network (FCN) on each individual actor's RoI, keeping the same perception network, but making separate predictions for each actor.
However, within the RoI of a pedestrian there might be other actors, and we observe that \emph{FCN} outputs spurious occupancy at those locations due to the translation invariance of convolutions, which translates into poor performance when recognizing other pedestrians.
To remove the translation invariance, we add CoordConv \cite{liu2018intriguing} to the RoI features of each actor, which improves all metrics (\emph{FCN+CoordConv}).
Next, we introduce actor-actor interactions in what we refer to as \emph{SA-GNN w/o scene}. This is essentially our model without the scene node. We can see that explicitly reasoning about interactions helps particularly in instance-level metrics.
Next, we add the scene node with actor-to-scene messages, but without scene-to-actor messages (\emph{SA-GNN w/o s2a}), which outperforms the scene-level metrics by a big margin but harms motion forecasting. 
Finally, we add the scene-actor messages to recover our full model, which achieves a better tradeoff between scene occupancy and motion forecasting performance. 
Moreover, including scene-to-actor messages adds robustness to missing detections. An example is collision avoidance with undetected pedestrians: the average occupancy probability placed by the motion forecasts of detected pedestrians on top of the ground-truth trajectories of undetected pedestrians reduces from 9.25\% to 5.46\%.

\begin{figure}[h]
    \centering
    \begin{tabular} {@{}c@{\hspace{.5em}}c@{\hspace{.5em}}c}
        {} & \textbf{Actor Motion} & \textbf{Scene Occupancy} \\
        \rotatebox[origin=c]{90}{\textbf{Scenario 1}} & \raisebox{-0.5\height}{\includegraphics[width=0.47\columnwidth, trim={1.25cm, 3.0cm, 1.5cm, 1.5cm}, clip]{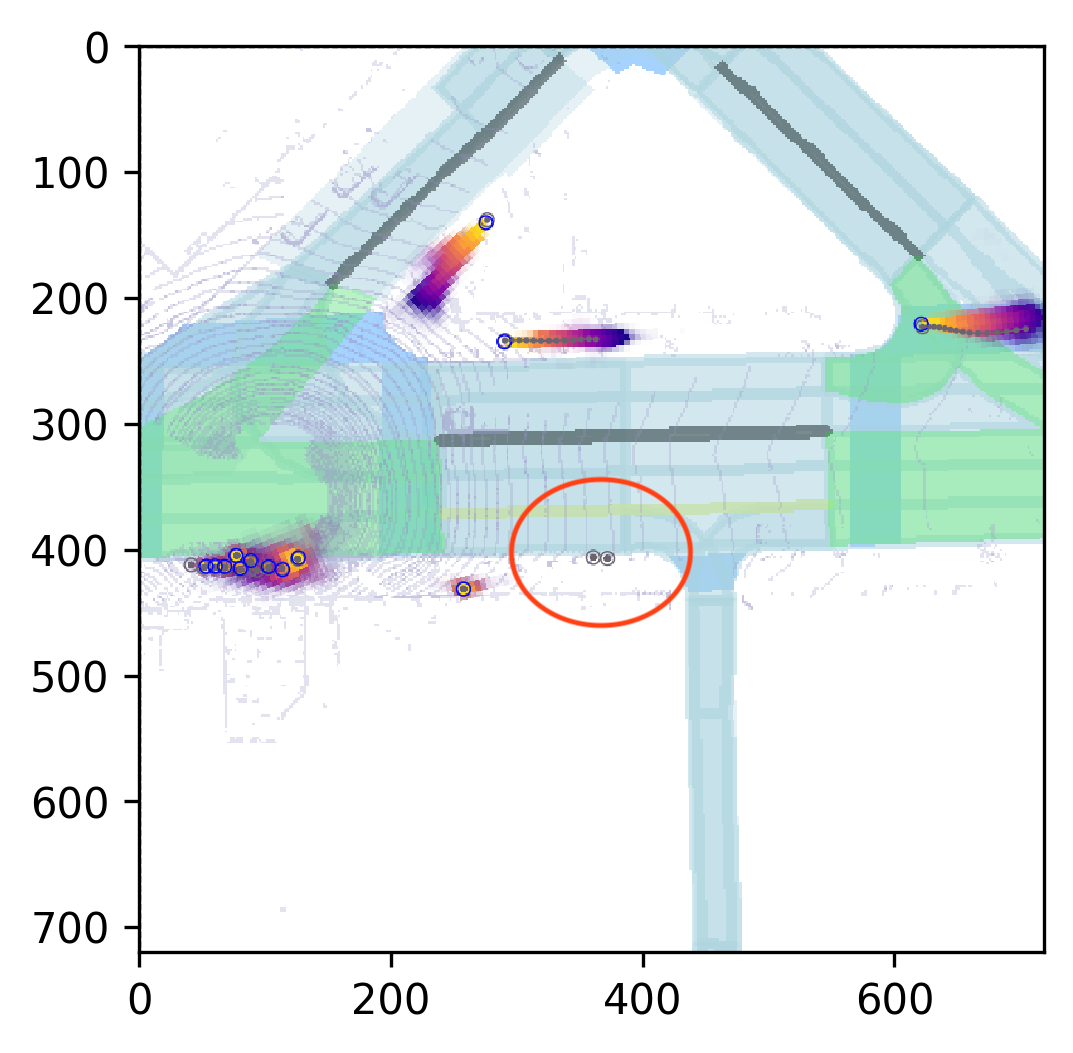}} & \raisebox{-0.5\height}{\includegraphics[width=0.47\columnwidth, trim={1.25cm, 3.0cm, 1.5cm, 1.5cm}, clip]{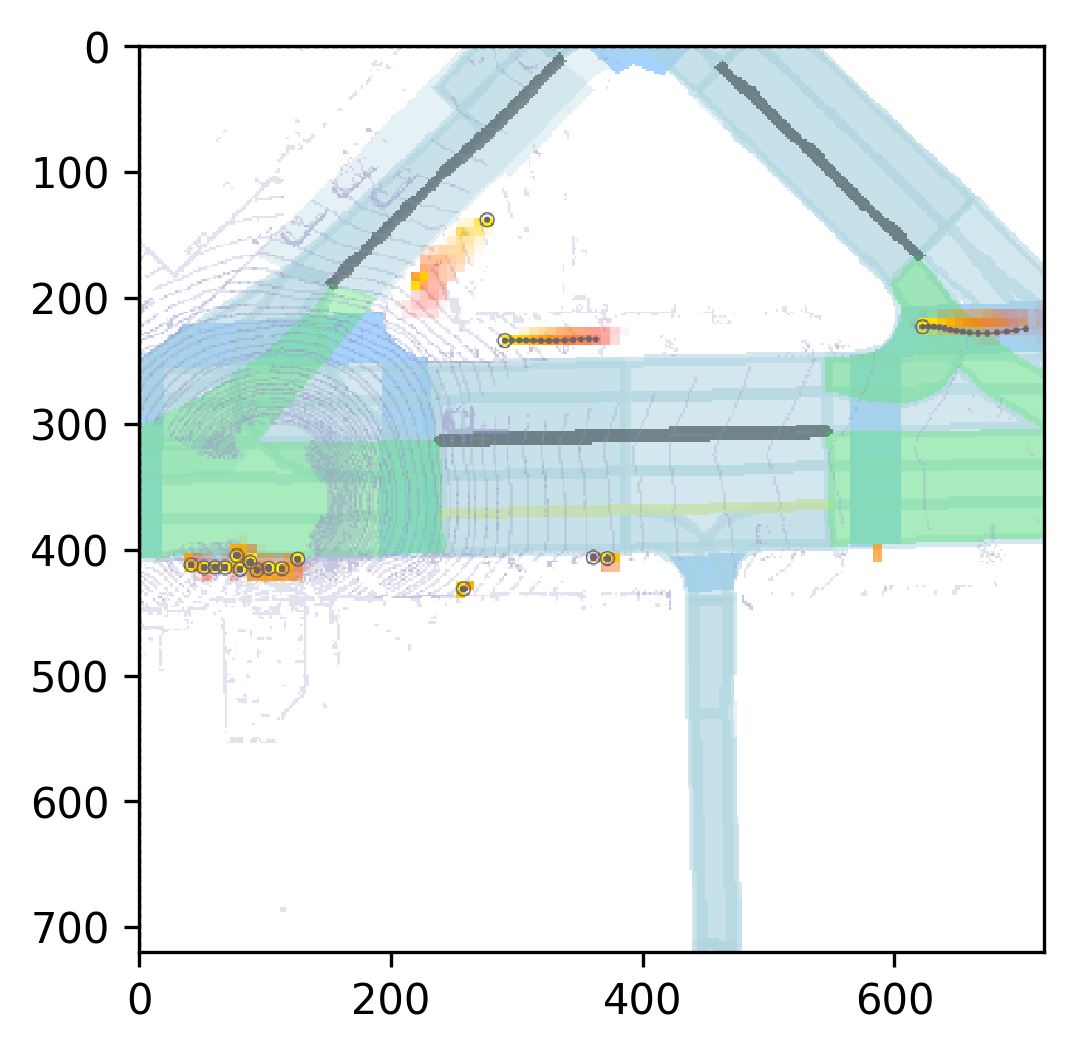}} \vspace{.5em} \\
        \rotatebox[origin=c]{90}{\textbf{Scenario 2}} & \raisebox{-0.5\height}{\includegraphics[width=0.47\columnwidth, trim={1.25cm, 3.0cm, 1.5cm, 1.5cm}, clip]{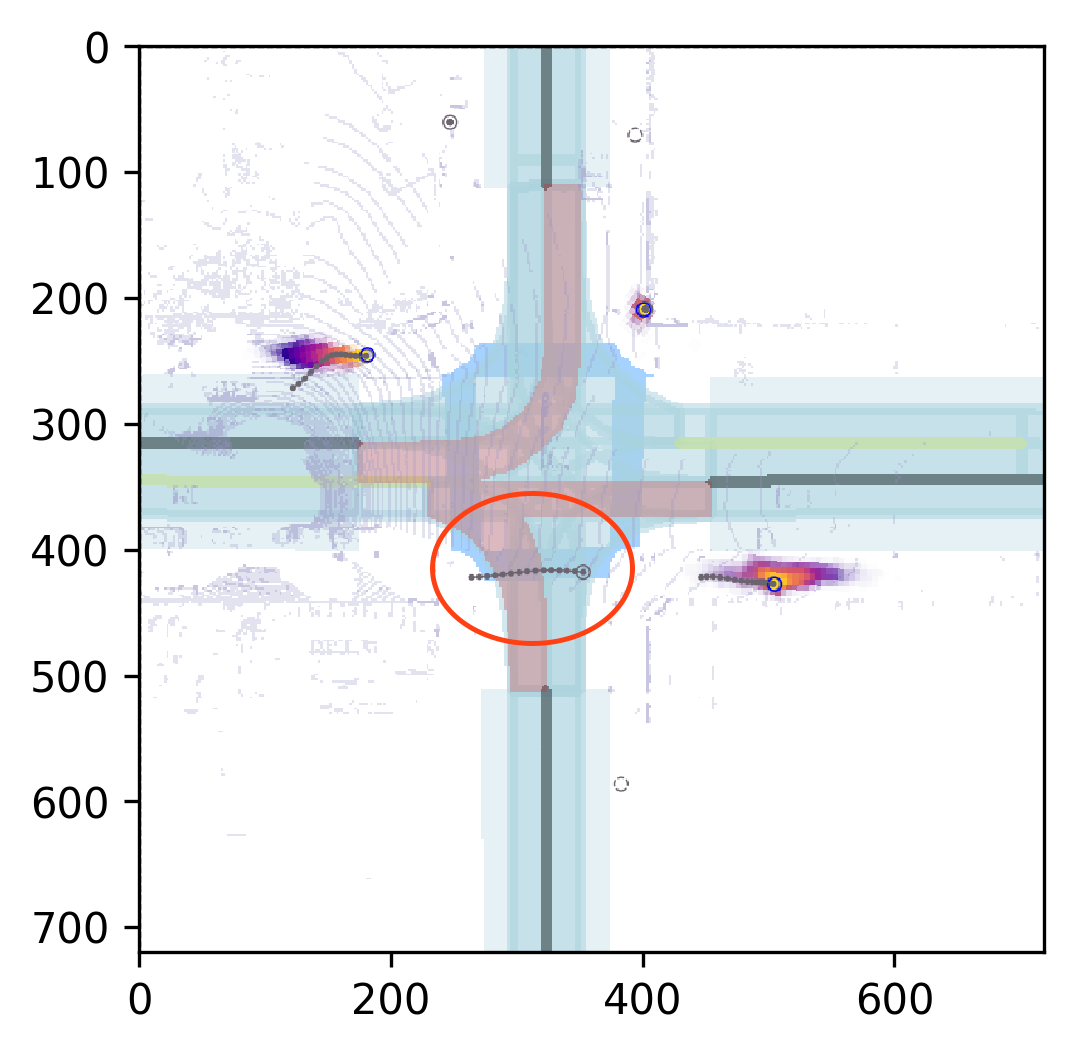}} & 
        \raisebox{-0.5\height}{\includegraphics[width=0.47\columnwidth, trim={1.25cm, 3.0cm, 1.5cm, 1.5cm}, clip]{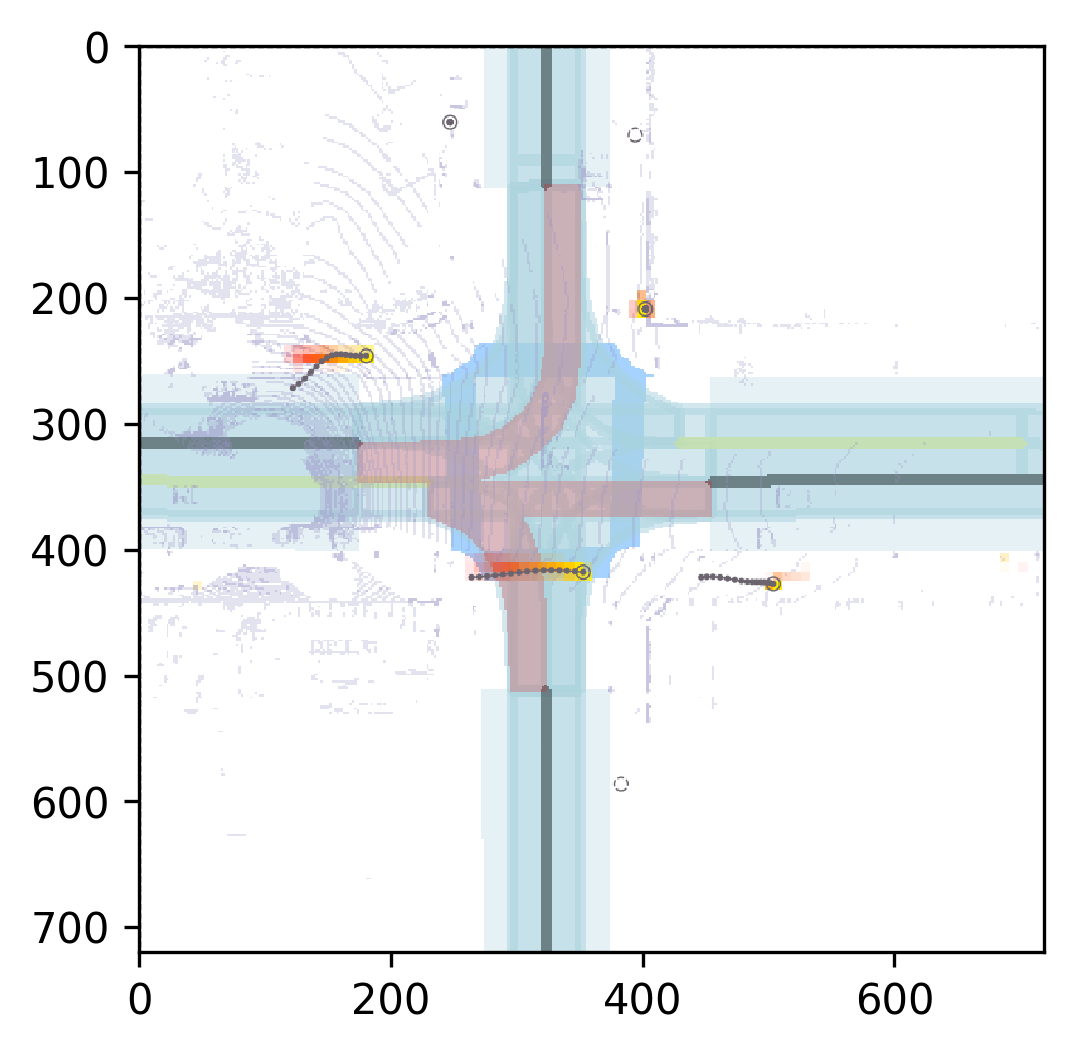}} \vspace{.5em} \\
        \rotatebox[origin=c]{90}{\textbf{Scenario 3}} & \raisebox{-0.5\height}{\includegraphics[width=0.47\columnwidth, trim={1.25cm, 3.0cm, 1.5cm, 1.5cm}, clip]{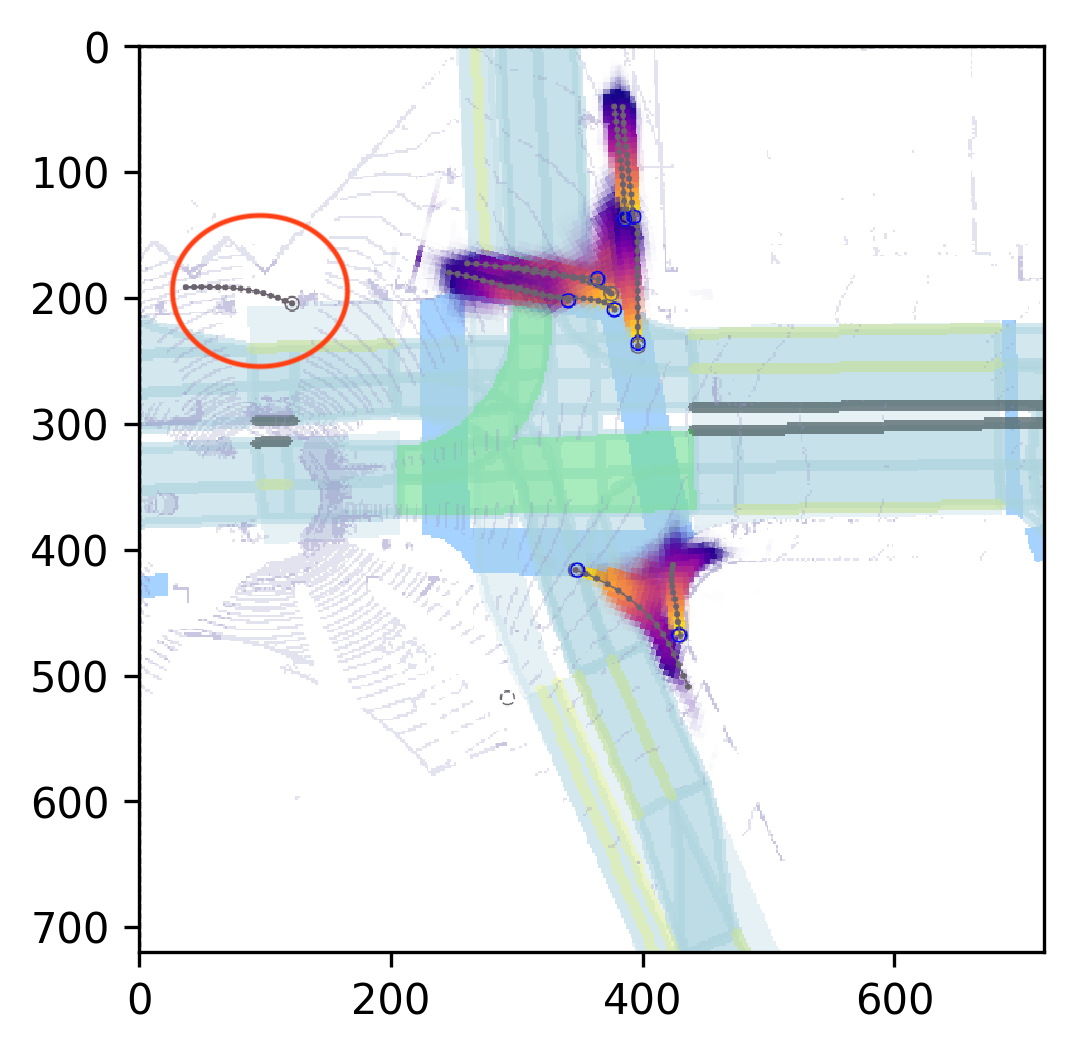}} \vspace{.5em} &
        \raisebox{-0.5\height}{\includegraphics[width=0.47\columnwidth, trim={1.25cm, 3.0cm, 1.5cm, 1.5cm}, clip]{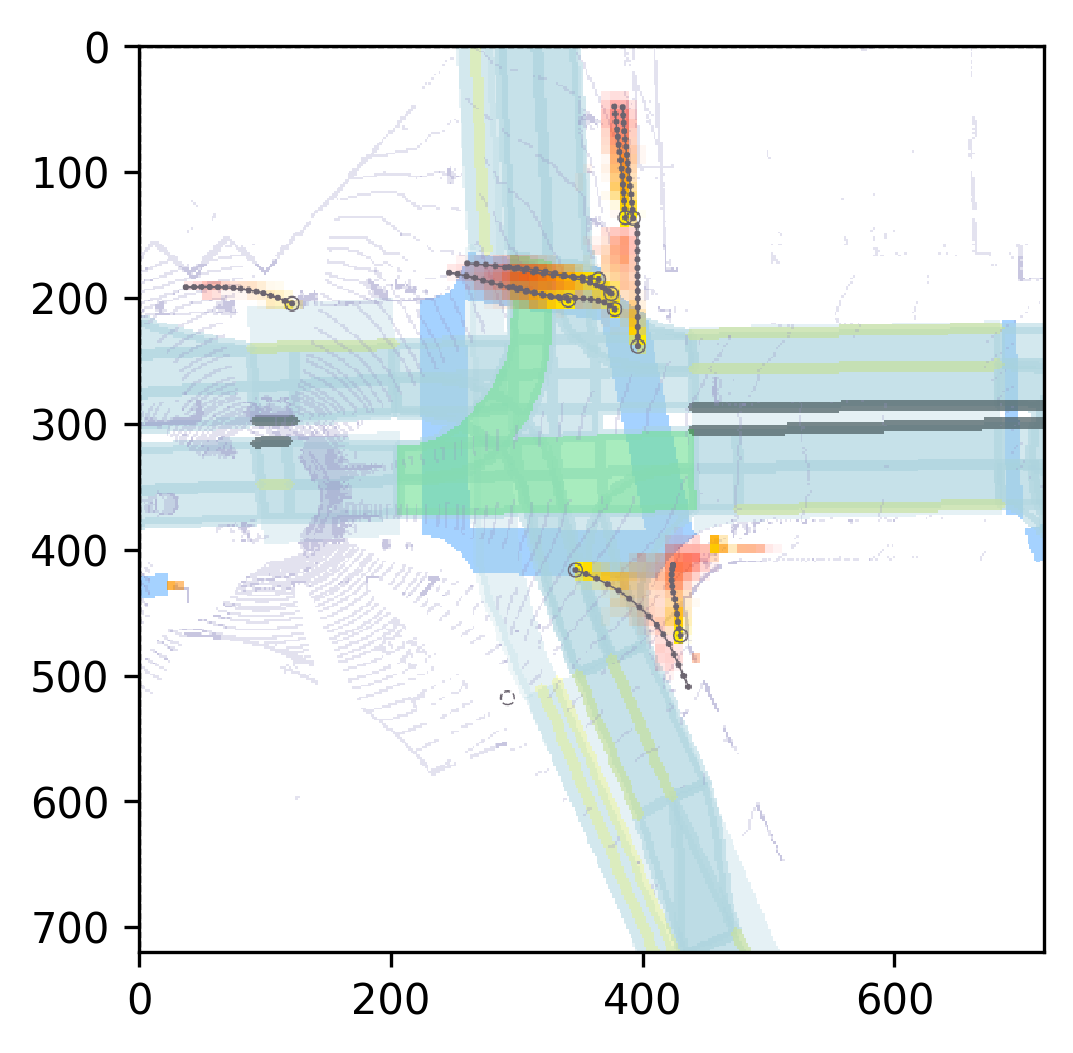}} \vspace{.5em}
    \end{tabular}
    \cutcaptionup
    \caption{\textbf{[\ourdataset{}] Qualitative results}. Ground-truth future trajectory are shown in gray. Detections are shown in blue in the first row, and red ellipses highlight false negatives recovered by the scene occupancy. The heatmaps' color indicates future horizon and transparency the probability.
    See more qualitative results in our \suppref.
    }
    \cutcaptiondown
    \label{fig:qualitative}
\end{figure}

\textbf{Qualitative Results:}
Fig.~\ref{fig:qualitative} shows examples of the predicted actor motion and scene occupancy forecasts from our model. Our object detector is operated at high precision (90 \%), and our scene occupancy forecasting guarantees high recall.

\cutsectionup
\section{Conclusion}
\cutsectiondown

In this paper, we build a {\ourmodel} ({\ourmodelshort}) to predict the pedestrians' motion and their shared occupancy map in the scene.
Our model captures the interaction among multiple pedestrians and takes the scene-level occupancy information into consideration so that it is aware of the missing detections.
We show the effectiveness of our method on two large-scale real-world datasets, a new dataset collected by ourselves and a publicly available dataset nuScenes.
Experimental results indicate that our {\ourmodelshort} achieves state-of-the-art performance, especially on the safety-oriented scene occupancy metrics.
Integration with a motion planner is left for future work.

\pagebreak
\bibliography{reference}

\clearpage
\appendices
\section{Implementation Details}

\textbf{LiDAR Voxelization:}
In \nuscenes, following \cite{zhu2019classbalanced} we set the voxel sizes to be $\Delta L = \Delta W = 0.1$ and $\Delta H = 0.2$ meters/pixel, and voxelize a region of 80 by 80 meters centered around the self-driving vehicle (SDV).
In \ourdataset, we follow \cite{yang2018pixor} and set $\Delta L = \Delta W = \Delta H = 0.2$ meters/pixel. We also voxelize an 80 by 80 meter region, but prioritizing the front of the vehicle with 70 meters, leaving only 10 meters behind, as the front is more important for safety and the forward speeds are much higher than backwards.
In our dataset \ourdataset, we also exploit the ground height information available in our HD maps to obtain a ground-relative height \cite{yang2018hdnet} instead of directly using the height returned by the LiDAR returns. 
In both datasets we use $N_\text{sweep}=10$ LiDAR sweeps (at 10 Hz).

\textbf{Backbone Architecture:}
We use a lightweight backbone network adapted from \cite{yang2018pixor} for efficient feature extraction and object detection.
In particular, we instantiate two separate branches such that the voxelized LiDAR and rasterized map are processed as two separate streams first, and fused later.
The resulting features from both branches are then concatenated along the feature dimensions since they share the same spatial resolution, and finally fused by a convolutional header. Our LiDAR backbone uses 2, 2, 3, and 6 layers in its 4 residual blocks. The convolutions in the residual blocks of our LiDAR backbone have 32, 64, 128 and 256 filters with a stride of 1, 2, 2, 2 respectively. The stream that processes the high-definition maps uses 2, 2, 3, and 3 layers in its 4 residual blocks. 
The convolutions in the residual blocks of our map backbone have 16, 32, 64 and 128 filters with a stride of 1, 2, 2, 2 respectively. 
For both streams of the backbone, the final feature map is a multi-resolution concatenation of the outputs of each residual block, as explained in~\cite{engelcke2017vote3deep}. This gives us 4x down-sampled features with respect to the input. 
The feature maps output by the first and second residual blocks are averaged pooled to this resolution, the third one is at this resolution, and the fourth one is interpolated.
The header network consists of 4 convolution layers with 256 filters per layer. The final scene context $\mathcal{C}$ is of size 256 x 100 x 100 in \ourdataset and 256 x 200 x 200 for \nuscenes, as we start with double the spatial resolution in the latter and we use the exact same network architecture for both datasets.
We use GroupNorm~\cite{wu2018group} because of our small batch size (number of frames) per GPU.

\textbf{Detection Header:}
We use two convolutional layers to output a classification (i.e. confidence) score, centroid and heading regression of the pedestrian for each anchor location following the output parameterization proposed in~\cite{yang2018pixor}, which are finally reduced to the final set of candidates by applying non-maximal suppression (NMS) with a distance of 0.3 meters, and finally thresholding low-probability detections at the detection threshold given by the desired common recall for evaluation.
Given our voxelization and backbone network architecture, there are 10 thousand anchors in \ourdataset and 40 thousand in \nuscenes. They are both at 4x downsampling from the original spatial size of the input, which means there is an anchor every 0.8 meters in \ourdataset and an anchor every 0.4 meters for \nuscenes.

\textbf{Output Parameterization Details:}
In both datasets, we employ a $64 \times 64$ spatial grid with a resolution of 0.5 meters/pixel for the actor motion parameterization, which is a hyperparameter to the RoiAlign operator.
For the scene occupancy we use a resolution of 0.8 m/pixel in \ourdataset and 0.4 m/pixel in \nuscenes, due to the differences in LiDAR voxelization for both datasets, as we keep the same architecture for both.

\textbf{Actor-to-actor Message Passing Architecture:} Here we explain how the vector messages and updated actor node representations are computed.

\cutsubsectionup
\begin{itemize}[leftmargin=*]
    \item \emph{Actor-to-actor vector messages.} %
    To reduce the memory consumption of our model, we design an actor-to-actor message function $F_{\text{actor}}$ that reduces an spatial feature map of size $D \times H_m \times W_m$ to a feature vector $D' \times 1 \times 1$, where $D=192, H_m = 64, W_m=64, D' = 256$. This way the message computation still leverages spatial convolutions as the node states are spatial feature maps, but the resulting messages are compressed to feature vectors. 
    $F_{\text{actor}}$ is a 3-layer fully convolutional network with kernels of size 8, 8, 3, strides of 4, 4, 1 and paddings of 2, 0, 0 respectively.
    In practice, we found that the performance of the model with learned vector messages is comparable to the performance of the model where messages keep their spatial dimensions.
    $F_{\text{embed}}$, the function that embeds each of the node representations before applying the message function is a 2-layer CNN.
    \item \emph{Actor state update.} First all the neighbors' messages are aggregated via max-pooling. 
    $F_{\text{deconv}}$ then brings the aggregated message back to spatial dimensions using transposed convolutions with the same layer parameters as described for $F_{\text{actor}}$ above.
    Finally, we update our model using the Convolutional Gated Recurrent Units (ConvGRU)~\cite{siam2017convolutional}, where the hidden state is the node's previous state, and the input is the deconvolved, aggregated message.
\end{itemize}
\cutsubsectiondown

\textbf{Actor-to-scene Message Passing Architecture:}
Here we explain how the actor-to-scene messages are computed and how the scene node representations are updated. 
We first apply $F_{\text{instance}}$ directly to each actor node state in the actor coordinate system to get the message in the actor coordinate frame. 
$F_{\text{instance}}$ is a 2-layer CNN with kernel size of 3 and padding of 1 that preserves the spatial dimensions.
Then, each actor node message gets bilinearly sampled into the scene coordinate frame ($\mathcal{T}_{I_i \rightarrow S}$ in the algorithm box in the main paper).
After the messages from all nodes are in the scene coordinate frame, we apply feature-wise max pooling to aggregate them.
Finally, another ConvGRU updates the scene node states.

\textbf{Scene-to-actor Message Passing Architecture:}
The scene node representation might contain information about undetected actors, and thus we want to leverage this to model potential interactions with the detected node's future motions. 
To compute the scene-to-actor messages, we first pool the scene node features at the associated rotated region of interest of each actor using Rotate RoI Align, then concatenate with the actor's previous state, and apply $F_{\text{scene}}$ to compute the message.
$F_{\text{scene}}$ has the same architecture as $F_{\text{actor}}$.

\cutsubsectionup
\section{Loss and Training Details}
\cutsubsectiondown

\textbf{Multi-task Learning Objective:}
We minimize a multi-task learning objective consisting of a binary cross entropy loss for the classification branch of the detection network, a regression smooth $\ell_1$ loss to fit detection bounding boxes, the negative log-likelihood for the spatial probability map prediction for each actor and a binary cross entropy per pixel for the scene occupancy.

The classification loss is as follows.
\cutequationup
\begin{equation}\mathcal{L}_\text{class} = - \sum_{i \in \mathcal{S}} \hat{c}_{i} \log (c_i) + (1 - \hat{c}_{i} ) \log (1 - c_i)\end{equation}
\cutequationdown
Here, $c_i$ is the confidence predicted for anchor i and $\hat{c}_{i}$ is 1 for positive anchors (pedestrian) and 0 for negative (background). $\mathcal{S}$ is the set of anchors selected for the loss. It is composed by all the positive anchors associated to pedestrians as well as a set of mined hard negatives~\cite{shrivastava2016training}.  Specifically, we use negative-positive ratio of 3-to-1.

The regression loss is as follows.
\cutequationup
\begin{equation}
\mathcal{L}_\text{reg} = - \sum_{i \in \mathcal{A}}  l(\Delta x_i) + l(\Delta y_i) + l(\sin\beta_{i}) + l(\cos\beta_{i})
\end{equation} 
\cutequationdown
Here, $\mathcal{A}$ is the set of positive anchors, $\Delta x_i$ and $\Delta y_i$ are the (x,y) offsets from the anchors, and $\beta_{i}$ is the predicted heading. 
$\ell(\cdot)$ is the smooth L1 loss or Huber loss.
\cutequationup
\begin{equation}l(m)=
        \begin{cases}
        0.5 \cdot m^2, \text{if} |m| < 1\\
        |m| - 0.5, \text{otherwise}
        \end{cases}\end{equation}
\cutequationdown

The actor motion loss is the negative log-likelihood of independent categorical distributions $I_{i}^{t}$ over time ($T_f$ time steps) and actors.
\cutequationup
\begin{equation}
\mathcal{L}_\text{pred} = - \sum_{i \in \mathcal{A}} \sum_{t=1}^{T_f} \log p_\theta (I_{i}^{t} \given \mathcal{C})
\end{equation}
\cutequationdown

For our scene header, we minimize a binary cross-entropy loss per spatial pixel and time, between our predicted occupancy probability map and their ground truth locations in the entire scene.
\cutequationup
\begin{equation}
    \mathcal{L}_{\text{Scene}} = -  \sum_{t=1}^{T_f} \sum_{i \in \mathcal{S}_t} \hat{S}_{i}^{t} \log p (S_{i}^{t} \given \mathcal{C}) + (1 - \hat{S}_{i}^{t}) \log (1 - p (S_{i}^{t} \given \mathcal{C}) )
\end{equation}
\cutequationdown
Here, $\mathcal{S}_t$ is the set of spatial pixels selected for loss at future time step $t$ and $\hat{S}_{i}^{t}$ is 1 if any pedestrian occupies location $i$ at future time step $t$ and 0 otherwise. We apply the same hard negative mining scheme as for the detection loss at each future time step.

The final loss is a linear combination of above losses, where $\lambda_1, \lambda_2, \lambda_3, \lambda_4$ are hyperparameters.
\cutequationup
\begin{equation}
\mathcal{L}_{\text{Inst}} = \lambda_1 \mathcal{L}_\text{class} + \lambda_2 \mathcal{L}_\text{reg} + \lambda_3 \mathcal{L}_\text{pred} + \lambda_4 \mathcal{L}_{\text{Scene}}    
\end{equation}
\cutequationdown
In practice we use $\lambda_1=1.0, \lambda_2=1.0, \lambda_3=1.0, \lambda_4=0.5$

\textbf{Scheduled Sampling:} For the actor-level motion forecasting training, we use scheduled sampling between ground-truth bounding boxes and detected bounding boxes.
This is important for 2 reasons. First, our predictions depend heavily on the quality of the detection outputs due to RoI Align for per-actor feature extraction. Thus, the features for prediction would be very noisy at the beginning of training if using the detected bounding boxes, since the whole model is trained from scratch. Second, our graph topology depends on the location of the actors since we use a k-nearest neighbors graph with limited ball distance radius. This means that when the detections show false positives/negatives, which they do since there is sensor noise and occlusion, there would be a mismatch in the distribution between the ground-truth topology and the one our model sees at inference when using detections.
In practice, we begin by using a 100\% ground truth pedestrian locations for 10-thousand iterations, and dropping by 10\% each subsequent 5-thousand iterations. Specifically, the percentage is applied at the frame level, not at the actor level.
The model is trained to convergence, after the warmup period, by using only detections. 

\textbf{Optimization:} In our experiments, we use Adam optimizer \cite{kingma2014adam} with a fixed learning rate $1.25 \times 10^{-5}$.
All models (including the baselines) were trained on 8 GPUs. Experiments were trained via distributed training with synchronized updates, with a batch size of 1 per GPU. We trained all models (including the baselines) from scratch, end-to-end.

\newcolumntype{s}{>{\centering\arraybackslash}X}
\begin{table*}[t]
	\centering
	\begin{tabularx}{\textwidth}{cccc|ss|sss|ss}
		\toprule
		
		\multicolumn{4}{c|}{\textbf{Multi-task coef.} } &  \multicolumn{2}{c|}{\textbf{Scene Occupancy} } &  \multicolumn{3}{c}{\textbf{Actor Motion Forecasting}} & \multicolumn{2}{c}{\textbf{Detection}} \\
		$\lambda_1$ & $\lambda_2$ &$\lambda_3$ &$\lambda_4$ &  Avg. mAP$\uparrow$  &  ACE$\downarrow$  &  NLL$\downarrow$ &  Exp. RMSE$\downarrow$ &  Argmax RMSE$\downarrow$ & mAP @0.3m & mAP @0.5m  \\
		\midrule
		1.0    &  1.0     &  1.0    &   1.0       & 37.72  &  4.29  &  \textbf{4.01}   &    2.34      &  \textbf{1.68}    &  73.57 & 81.55  \\
		1.0    &  1.0     &  0.5    &   1.0       & 37.52  &  3.00  &  4.06   &      2.38    &  1.74    & \textbf{74.58} & \textbf{81.88}  \\
		1.0    &  1.0     &  1.0    &   0.5       &  37.69      &    \textbf{1.21}  &    4.02&             2.33&           1.70      &  73.45 & 81.06  \\
		1.0    &  1.0     &  2.0    &   1.0       &  37.48 &  3.12  &  4.03  &      \textbf{2.30}    &  1.69    & 72.32  &  80.55 \\
		1.0    &  1.0     &  1.0    &   2.0       &  \textbf{38.53} &  2.26  &  4.02 &    2.35      &   1.70   & 73.47 & 81.11  \\
		\bottomrule
	\end{tabularx}
	\caption{\textbf{[\ourdataset] Ablation study} of the multi-task learning coefficients.}
	\cuttabledown
	\label{table:atg4d_ablation_lambdas}
\end{table*}

\begin{table*}[t]
	\centering
	\begin{tabularx}{\textwidth}{l|ssss|ssss}
		\toprule
		
		Model &  \multicolumn{4}{c|}{\textbf{Scene Occupancy} } &  \multicolumn{4}{c}{\textbf{Actor Motion Forecasting}} \\
		{} &  Avg. mAP$\uparrow$  &  Final mAP$\uparrow$ &  ACE$\downarrow$ & MCE$\downarrow$ &  NLL$\downarrow$ &  Exp. RMSE$\downarrow$ &  Argmax RMSE$\downarrow$ &  MinRMSE$\downarrow$ @25 \\
		\midrule
		\ourmodelshort w/o SS              &  22.98      &    10.70   &    28.96 &  57.22    &    4.42 &             2.72 &           1.88       &  0.80     \\
		\ourmodelshort (Ours)             &  \textbf{37.69}      &    \textbf{19.28}   &    \textbf{1.21}  &  \textbf{2.55}    &    \textbf{4.02} &             \textbf{2.33} &           \textbf{1.70}       &  \textbf{0.73}     \\
		\bottomrule
	\end{tabularx}
	\caption{\textbf{[\ourdataset{}] Ablation study} for the use of scheduled sampling (SS) of ground-truth vs. detection bounding boxes during training.}
	\cuttabledown
	\label{table:atg4d_ablation_scheduledsampling}
\end{table*}

\cutsubsectionup
\section{Baselines}
\cutsubsectiondown

Here we discuss (i) how we adapt the baselines to the joint perception and prediction setting, and (ii) how we evaluate the different types of outputs with a common set of metrics

\textbf{Joint Perception and Prediction Setting:}
STGAT~\cite{huang2019stgat},
SocialSTGCNN~\cite{mohamed2020social},
MultiPath~\cite{chai2019multipath},
and DRF-Net~\cite{jain2019discrete} were originally presented in the multi-agent trajectory prediction setting, where perception is assumed given and usually perfect (i.e. ground-truth past trajectories of each actor are used). 
However, this is not realistic in self-driving where perception is challenging and will have failures.
Thus, we adapt the baselines by replacing the actor context extracted from past trajectories on these methods by our per-actor feature extractor network from sensor data.
We note that we train the backbone network from scratch for each baseline and our model, and thus it is not biased; and SpAGNN~\cite{casas2019spatially}, which was proposed for the joint perception and prediction setting, showed that this adaptation yields better results that directly applying the methods proposed by the baselines on top of past trajectories extracted from previous object detections and an off-the-shelf tracker.

\textbf{Actor Motion Evaluation:}
Different methods have different output parameterizations of actor future motion. 
SpAGNN~\cite{casas2019spatially}, SocialSTGCNN \cite{mohamed2020social},
MultiPath~\cite{chai2019multipath} leverage simple output distributions such as bivariate gaussians to parameterize the output with a well-known, closed-form distribution at each future waypoint. 
STGAT~\cite{huang2019stgat} is a generative model which output are samples from the trajectory distribution.
Finally, DRF-Net~\cite{jain2019discrete} and our model \ourmodelshort output a categorical distribution over a spatial grid.
To evaluate all models in a common set of metrics we benchmark them in the spatial grid representation.
We do this as follows. 
For gaussian-based outputs with closed form probability density function (pdf), we integrate the pdf for each spatial grid cell to obtain the probability mass function (pmf), which is directly comparable to ours. 
Because integral methods are slow, we sample a denser grid than the one used to represent actor motion, evaluate the likelihood at each point, average the likelihood of the sampling points within a grid cell, and multiply by its area. We have checked and this approximation error is not significant. 
For sample-based approaches, we estimate the pdf by kernel density estimation (KDE) on 50 trajectory samples. We do this estimation for each time step independently. After we obtain a pdf, we use the same discretization method as described for Gaussian-based approaches to obtain the final pmf.

\textbf{Scene-occupancy Evaluation:}
To compare our scene-based predictions with existing baseline methods, we aggregate the baselines' individual predictions. 
Because of the assumption that actor motion forecasts are conditionally independent across actors given the scene context $\mathcal{C}$, we can aggregate them to get a probabilistic scene comparison for $P(\{S^t \vert t = 0, \cdots, T_f \} \vert \mathcal{C})$, for each scene spatial pixel $i$ in the scene grid.

\cutequationup
\begin{align}
    P(S_{i}^t \vert \mathcal{C}) &= P\big(\cup_n  (I_{n, i}^{t} ) \given \mathcal{C}\big) \nonumber\\
    &= 1 - P( \cap_n  I_{n,i}^{'t} \given \mathcal{C}\big)\nonumber\\
    &= 1 - \prod_{n}\big( 1 - P( I_{n,i}^{t}) \given \mathcal{C})\big)
\end{align}
\cutequationdown
Here, $I_{n,i}^t$ is the instance based prediction for actor $n$, at time step $t$, at the location of the $i$-th scene grid cell. $I_{n,i}^{'t}$ is its complement. This way we are able to compare the baselines to our scene-based predictions despite them not having such a predicted output. Because the locations of the grid cells on the actor coordinate frame and scene coordinate frame do not necessarily align, in practice we bilinearly interpolate $I_{n,i}^t$ (with re-normalization).

\section{Additional Results on \ourdataset}

\textbf{Multi-task Loss Coefficients Ablation:}
Table \ref{table:atg4d_ablation_lambdas} shows the results from different multi-task learning loss coefficients' configurations. 
The results show that
(i) our method is pretty robust to the multiple losses' coefficients since the changes are not very pronounced,
and (ii) there is no clear configuration for the hyperparameters as it poses a tradeoff between scene occupancy, actor motion forecasting and detection metrics.

\textbf{Scheduled Sampling Ablation:}
Table \ref{table:atg4d_ablation_scheduledsampling} shows a clear advantage for using scheduled sampling during training to transition between using ground-truth detections to train prediction and using the detections predicted by the model. This is very important because the object detections directly affect (i) the rotated RoI of each actor and (ii) the graph topology due to our k-nearest neighbor adjacency.

\begin{figure*}[h]
    \centering
    \begin{tabular} {@{}c@{\hspace{.5em}}c@{\hspace{.5em}}c@{\hspace{.5em}}c}
        \rotatebox[origin=c]{90}{\textbf{STGAT}} & \raisebox{-0.5\height}{\includegraphics[width=0.32\textwidth, trim={0.1cm, 5.0cm, 0.1cm, 3.0cm}, clip]{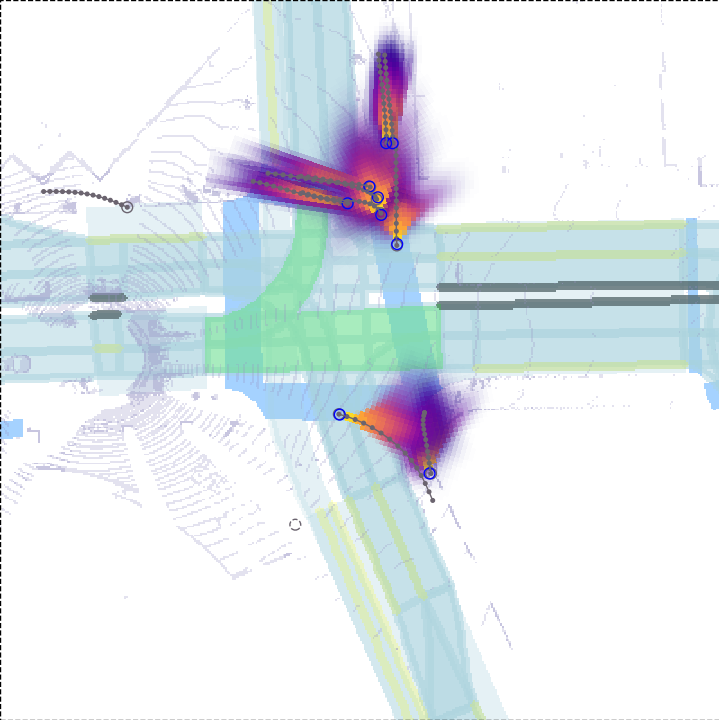}} &
        \raisebox{-0.5\height}{\includegraphics[width=0.32\textwidth, trim={0.1cm, 5.0cm, 0.1cm, 3.0cm}, clip]{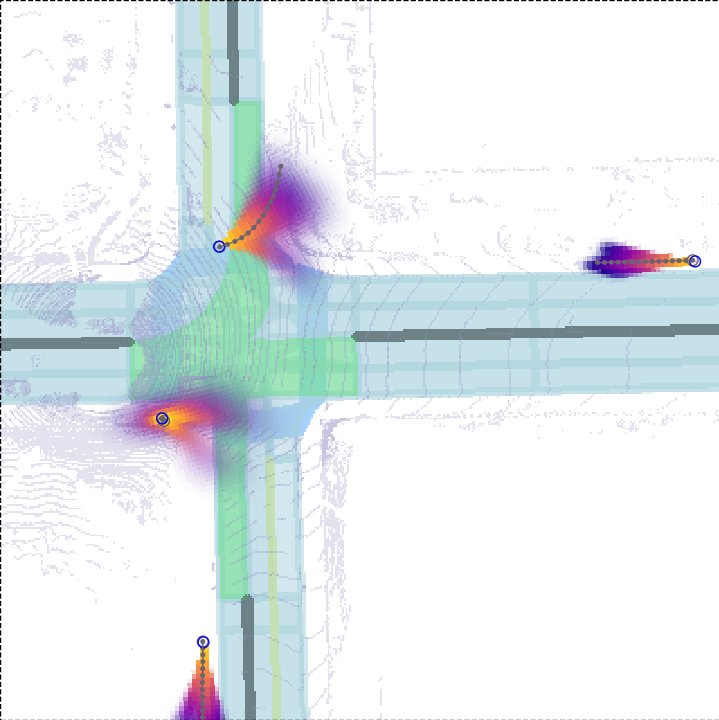}} & 
        \raisebox{-0.5\height}{\includegraphics[width=0.32\textwidth, trim={0.1cm, 5.0cm, 0.1cm, 3.0cm}, clip]{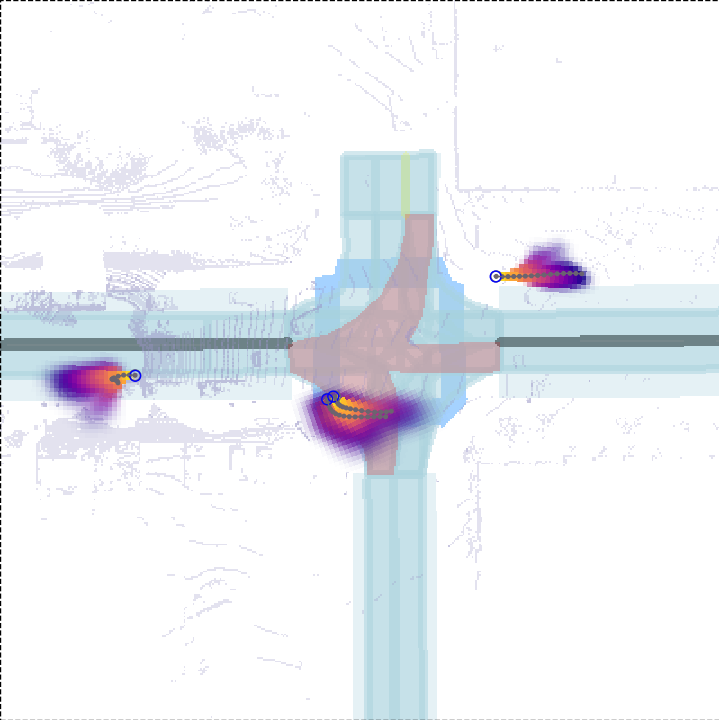}} \vspace{.5em} \\
        \midrule
        \rotatebox[origin=c]{90}{\textbf{SocialSTGCNN}} & \raisebox{-0.5\height}{\includegraphics[width=0.32\textwidth, trim={0.1cm, 5.0cm, 0.1cm, 3.0cm}, clip]{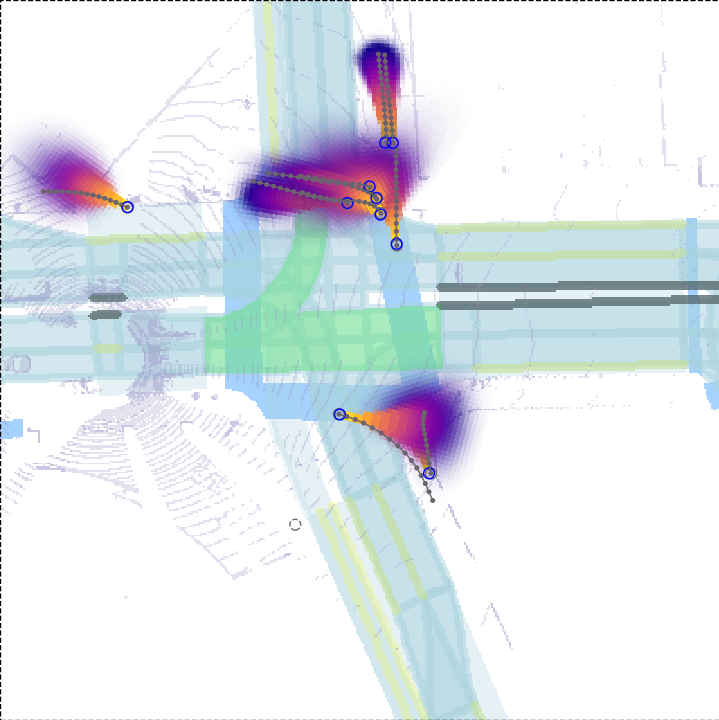}} &
        \raisebox{-0.5\height}{\includegraphics[width=0.32\textwidth, trim={0.1cm, 5.0cm, 0.1cm, 3.0cm}, clip]{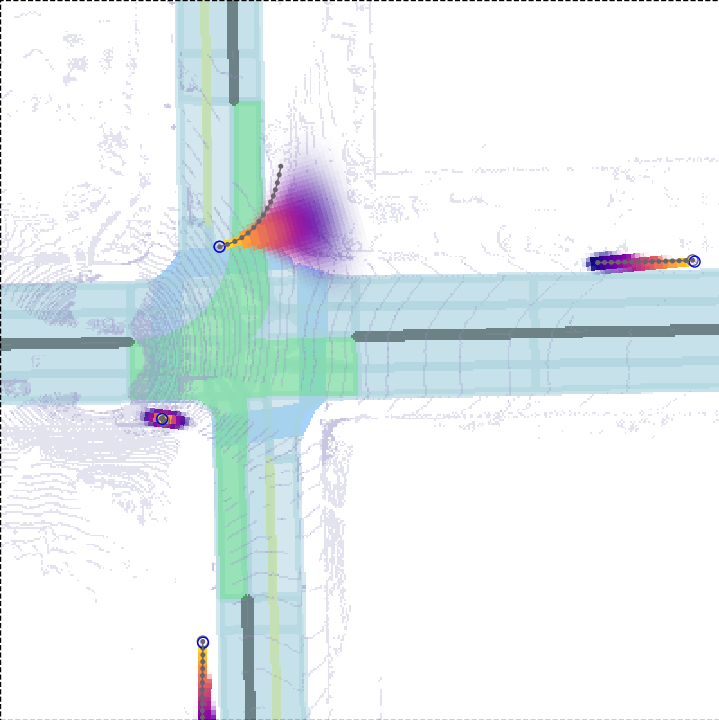}} &
        \raisebox{-0.5\height}{\includegraphics[width=0.32\textwidth, trim={0.1cm, 5.0cm, 0.1cm, 3.0cm}, clip]{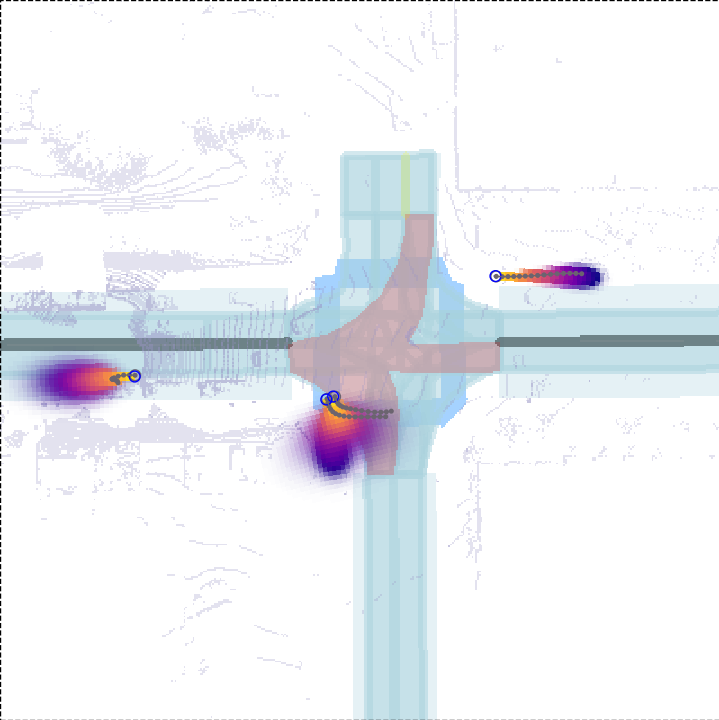}} \vspace{.5em} \\
        \midrule
        \rotatebox[origin=c]{90}{\textbf{SpAGNN}} & \raisebox{-0.5\height}{\includegraphics[width=0.32\textwidth, trim={0.1cm, 5.0cm, 0.1cm, 3.0cm}, clip]{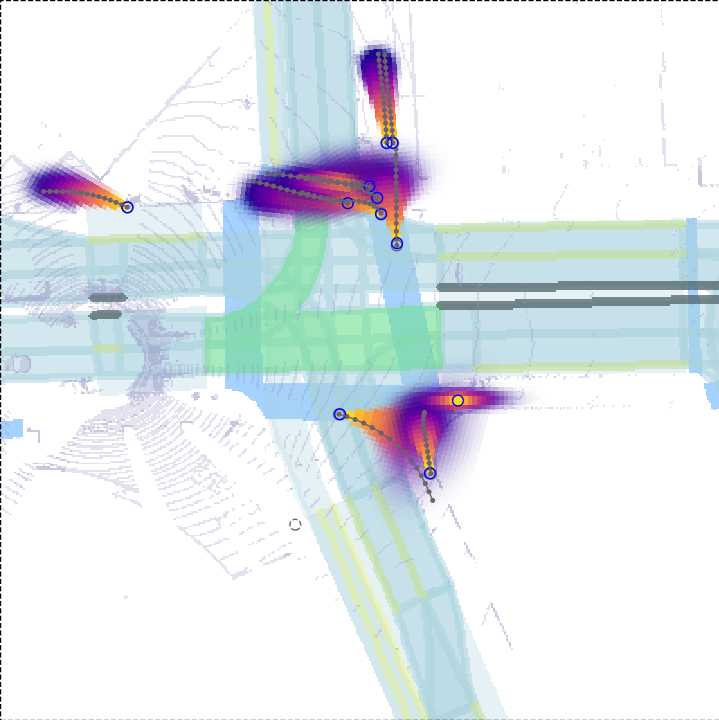}} &
        \raisebox{-0.5\height}{\includegraphics[width=0.32\textwidth, trim={0.1cm, 5.0cm, 0.1cm, 3.0cm}, clip]{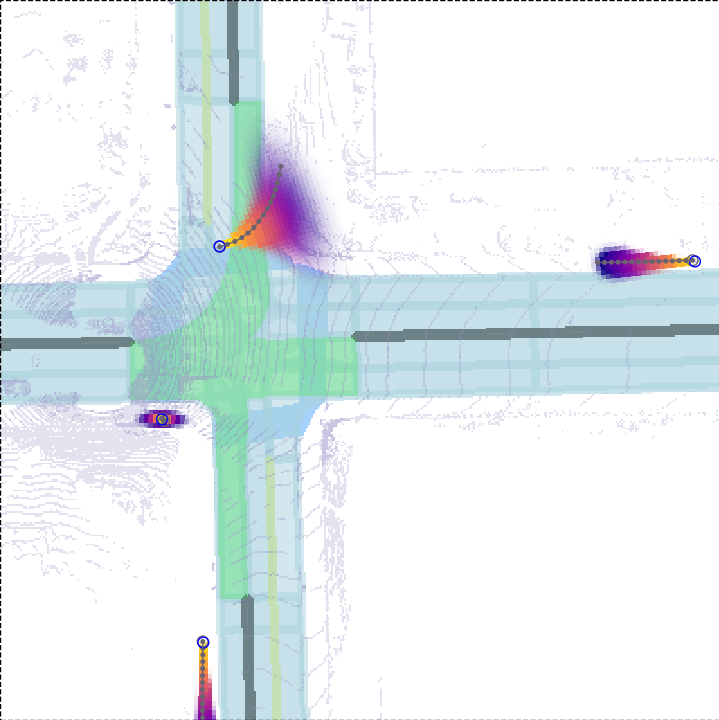}} & 
        \raisebox{-0.5\height}{\includegraphics[width=0.32\textwidth, trim={0.1cm, 5.0cm, 0.1cm, 3.0cm}, clip]{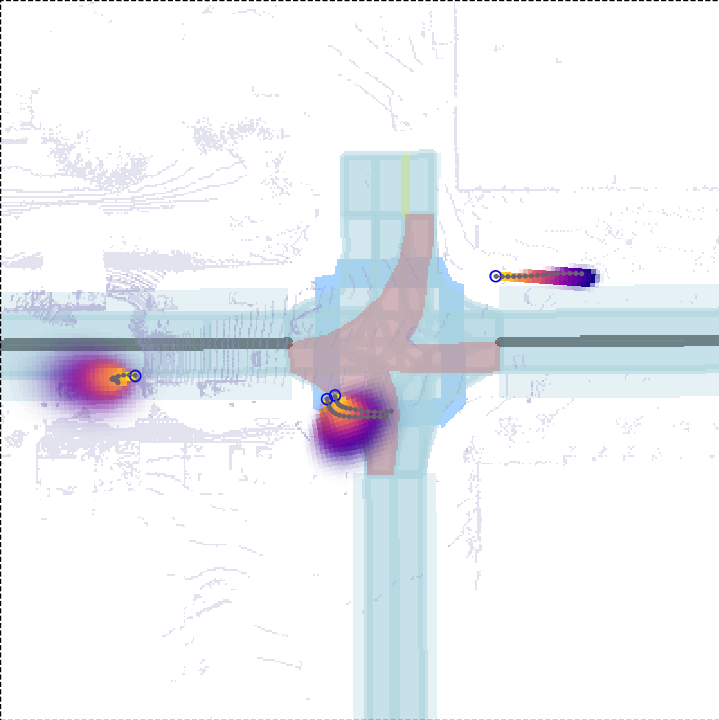}} \vspace{.5em} \\
        \midrule
        \rotatebox[origin=c]{90}{\textbf{MultiPath}} & \raisebox{-0.5\height}{\includegraphics[width=0.32\textwidth, trim={0.1cm, 5.0cm, 0.1cm, 3.0cm}, clip]{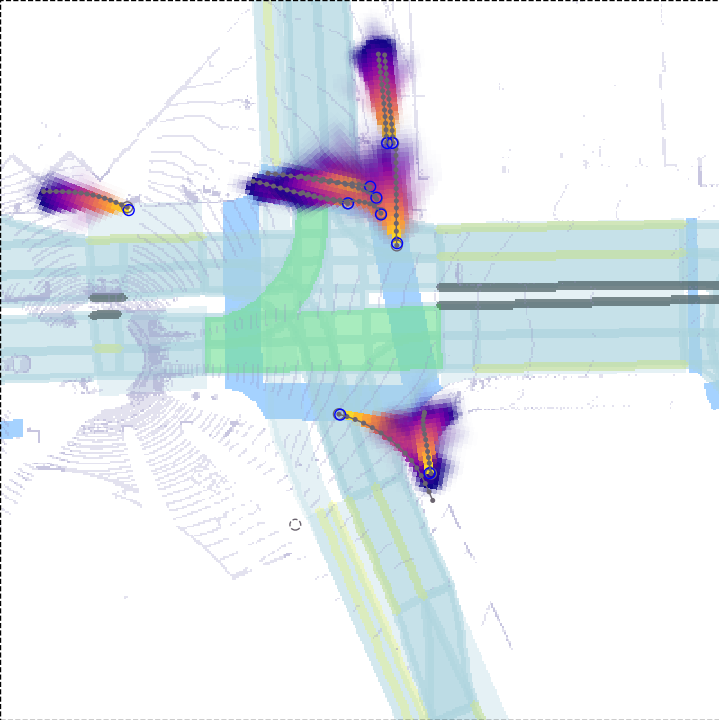}} &
        \raisebox{-0.5\height}{\includegraphics[width=0.32\textwidth, trim={0.1cm, 5.0cm, 0.1cm, 3.0cm}, clip]{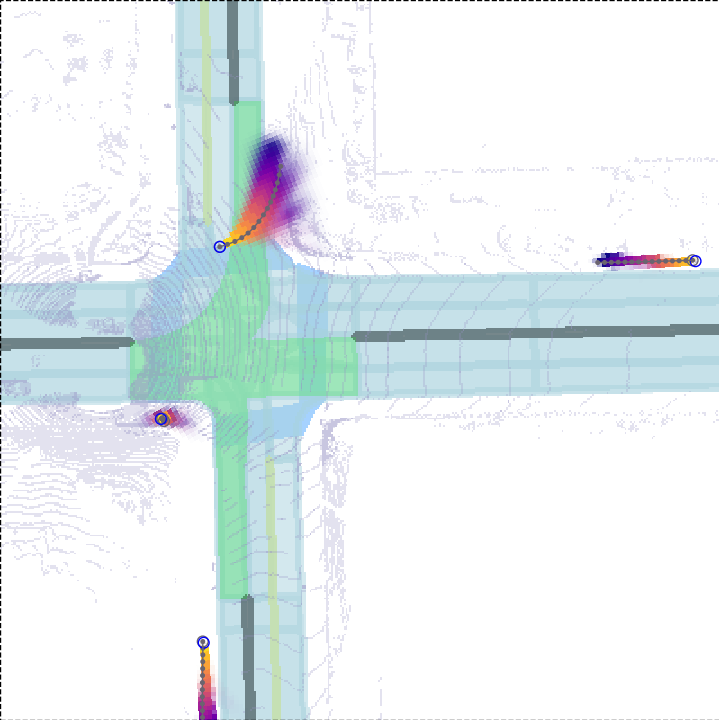}} & 
        \raisebox{-0.5\height}{\includegraphics[width=0.32\textwidth, trim={0.1cm, 5.0cm, 0.1cm, 3.0cm}, clip]{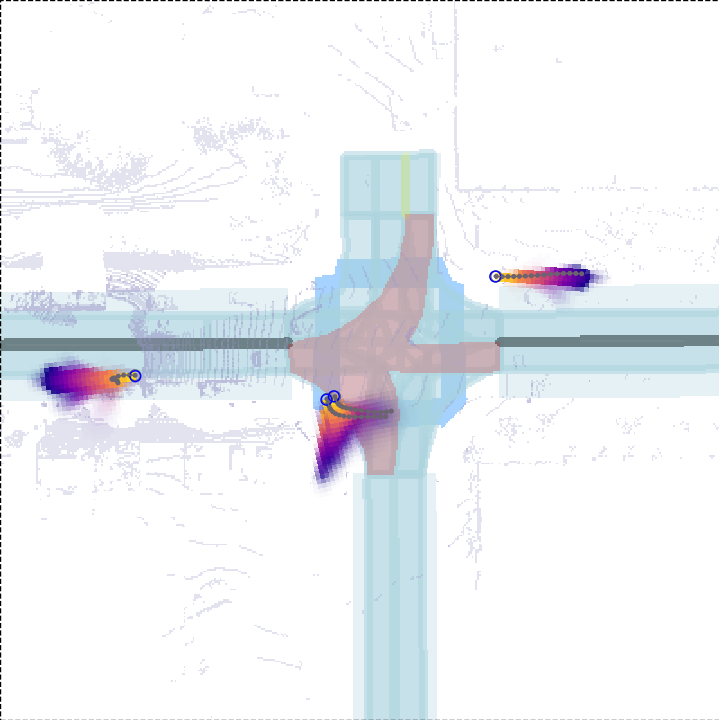}} \vspace{.5em} \\
        \midrule
        \rotatebox[origin=c]{90}{\textbf{DRF-Net}} & \raisebox{-0.5\height}{\includegraphics[width=0.32\textwidth, trim={0.1cm, 5.0cm, 0.1cm, 3.0cm}, clip]{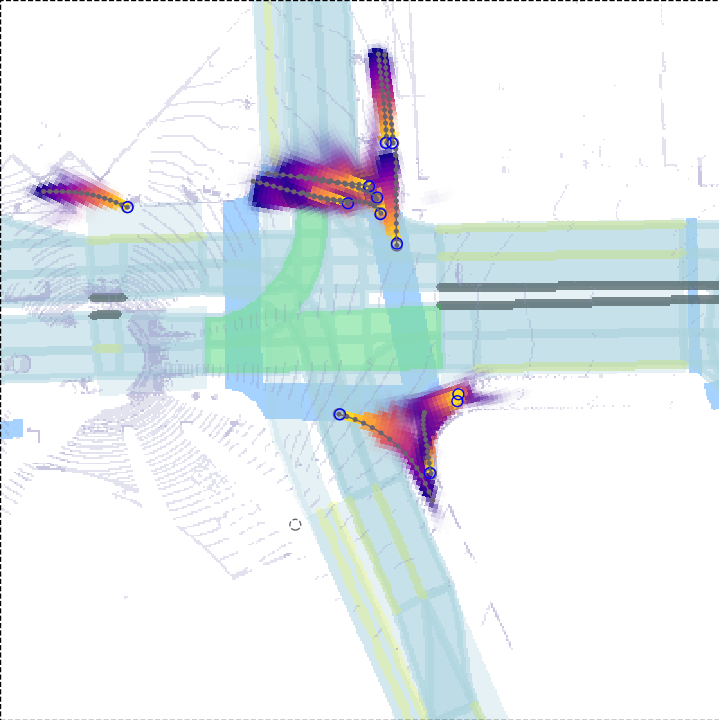}} &
        \raisebox{-0.5\height}{\includegraphics[width=0.32\textwidth, trim={0.1cm, 5.0cm, 0.1cm, 3.0cm}, clip]{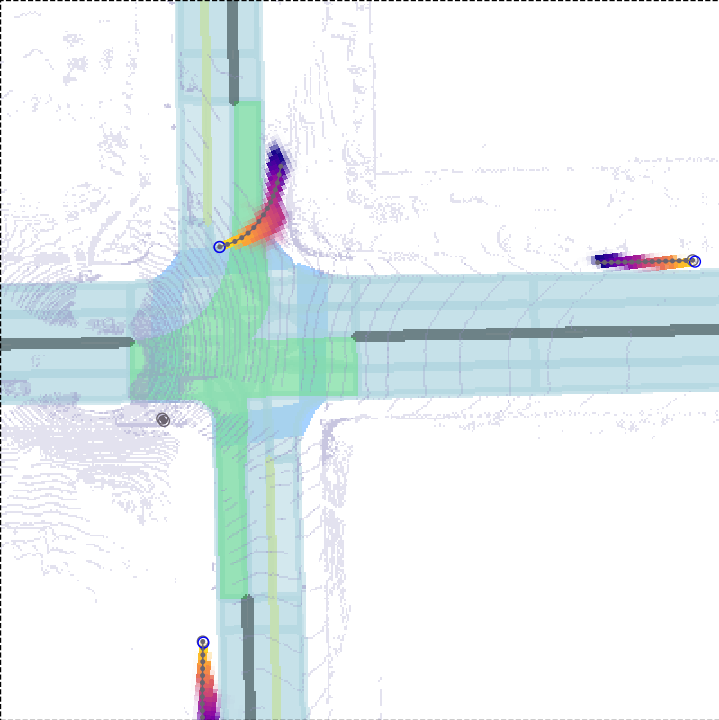}} & 
        \raisebox{-0.5\height}{\includegraphics[width=0.32\textwidth, trim={0.1cm, 5.0cm, 0.1cm, 3.0cm}, clip]{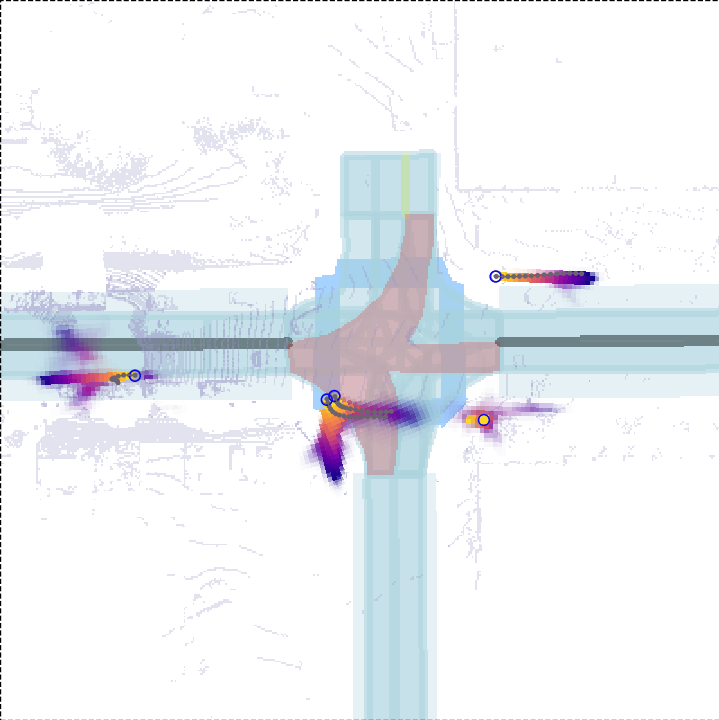}} \vspace{.5em} \\
        \midrule
        \rotatebox[origin=c]{90}{\textbf{SA-GNN (Ours)}} & \raisebox{-0.5\height}{\includegraphics[width=0.32\textwidth, trim={0.1cm, 5.0cm, 0.1cm, 3.0cm}, clip]{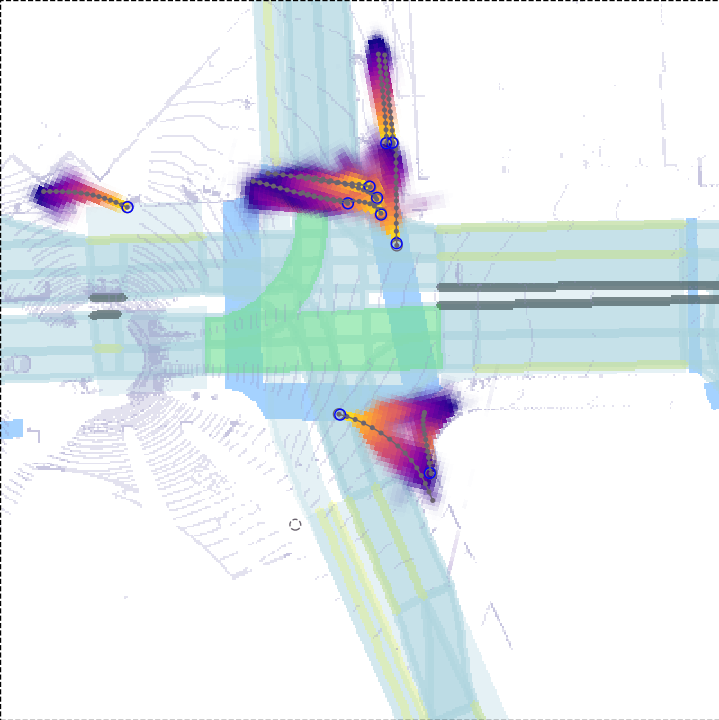}} &
        \raisebox{-0.5\height}{\includegraphics[width=0.32\textwidth, trim={0.1cm, 5.0cm, 0.1cm, 3.0cm}, clip]{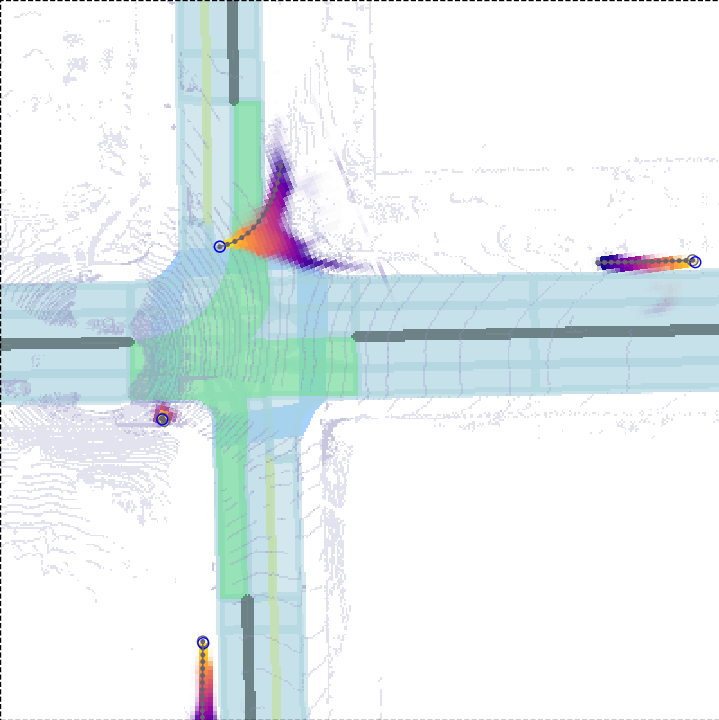}} & 
        \raisebox{-0.5\height}{\includegraphics[width=0.32\textwidth, trim={0.1cm, 5.0cm, 0.1cm, 3.0cm}, clip]{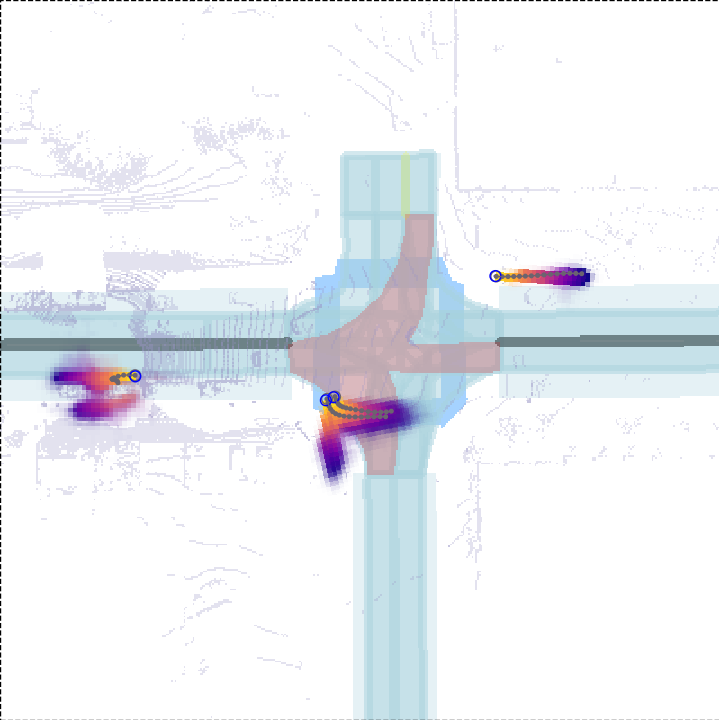}} \vspace{.5em} \\
    \end{tabular}
    \cutcaptionup
    \caption{\textbf{[\ourdataset{}] Qualitative motion forecasting comparison} with baselines. Left/Middle: \ourmodelshort predicts multimodal distributions with well-defined modes at intersections. Right: \ourmodelshort understands that probability mass should not be place on top of parked cars and predicts complex multimodal distributions to avoid such behavior.
    }
    \cutcaptiondown
    \label{fig:qualitative_supp_mf}
\end{figure*}

\textbf{Qualitative Comparison:}
Figures \ref{fig:qualitative_supp_mf} and \ref{fig:qualitative_supp_so} showcase qualitative comparisons between our \ourmodelshort and the baselines in the tasks of actor motion and scene occupancy forecasting, respectively.

\section{\nuscenes Results}

\newcolumntype{s}{>{\centering\arraybackslash}X}
\begin{table*}[t]
	\centering
	\begin{tabularx}{\textwidth}{l|ssss|ssss}
		\toprule
		Model &  \multicolumn{4}{c|}{\textbf{Scene Occupancy}} &  \multicolumn{4}{c}{\textbf{Actor Motion Forecasting}} \\
	    & Avg. mAP$\uparrow$  &  Final mAP$\uparrow$ &  ACE$\downarrow$ & MCE$\downarrow$ &  NLL$\downarrow$ &  Exp. RMSE$\downarrow$ &  Argmax RMSE$\downarrow$ &  MinRMSE$\downarrow$ @25 \\
		\midrule
		STGAT \cite{huang2019stgat}                     &   6.70       &   1.60     &  25.19     &  55.80     &  4.61    &    2.27           &   1.45    &   0.54     \\
		SocialSTGCNN \cite{mohamed2020social}           &   12.54      &   3.02     &  \textbf{17.52}     &  42.52     &  3.91    &    1.50           &   1.08    &   0.39     \\
		SpAGNN \cite{casas2019spatially}                &   11.21      &   2.77     &  23.22     &  42.63     &  4.06    &    \textbf{1.28}           &   \textbf{1.00}    &   \textbf{0.36}     \\
		MultiPath \cite{chai2019multipath}              &   10.48      &   2.99     &  24.98     &  42.28     &  \textbf{3.75}    &    \textbf{1.28}           &   1.01    &   0.37     \\
		DRF-Net \cite{jain2019discrete}                 &   11.76      &   3.17     &  25.21     &  57.07     &  3.89    &    1.30           &   1.09    &   \textbf{0.36}     \\
		\ourmodelshort (Ours)                                 &   \textbf{19.74}      &   \textbf{7.08}     &  19.09     &  \textbf{29.72}     &  3.77    &    1.33           &   1.05    &   0.37     \\
		\bottomrule
	\end{tabularx}
	\caption{\textbf{[\nuscenes{}] Scene occupancy and motion forecasting evaluation}.}
	\cuttabledown
	\label{table:results_nuscenes}
\end{table*}
\begin{figure}[t]
    \centering
    \begin{tabular} {@{\hspace{.1em}}c@{\hspace{.5em}}c}
        {\includegraphics[width=0.53\columnwidth]{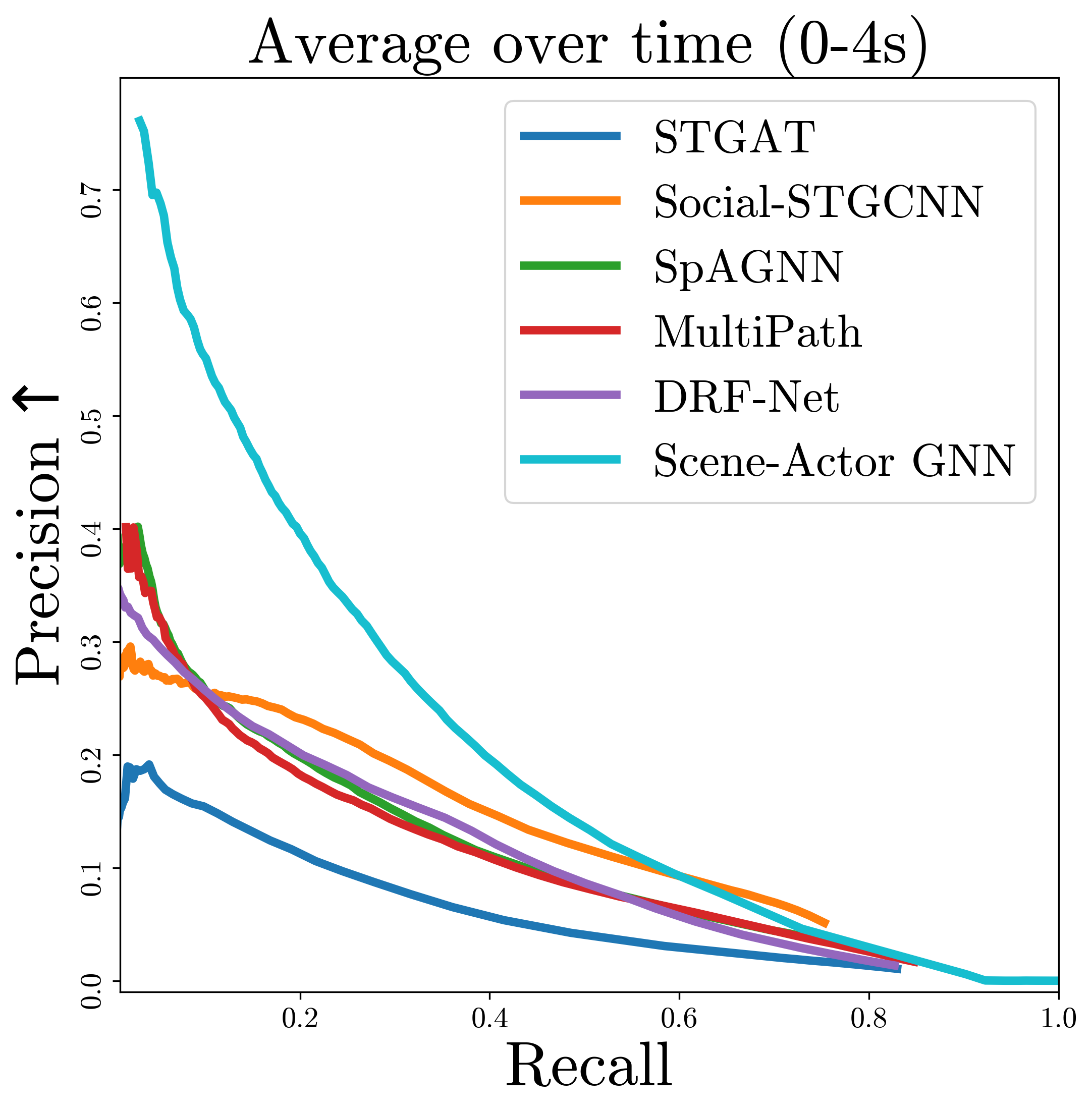}} &
        {\includegraphics[width=0.45\columnwidth]{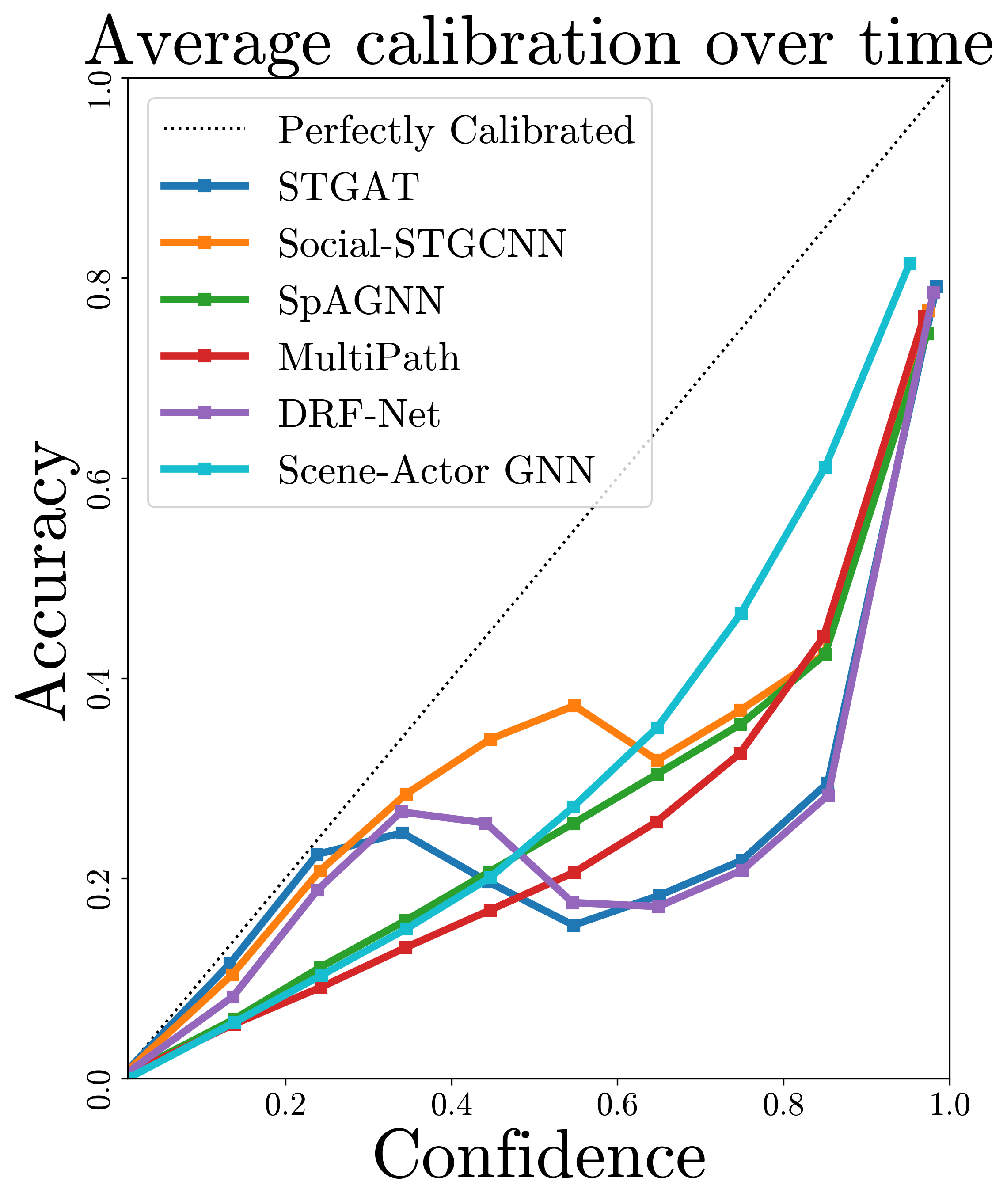}} \vspace{.5em}
    \end{tabular}
    \cutcaptionup
    \caption{\textbf{[\nuscenes{}] Scene occupancy evaluation}. Precision-Recall (PR) curves and reliability diagram}
    \cutcaptiondown
    \label{fig:scene_nuscenes}
\end{figure}

\begin{table}[t]
	\centering
	\begin{tabular}{c|cccc}
	\toprule
	\multirow{2}{*}{Model} & \multicolumn{4}{c}{Detection AP @ meter} \\
	~ & 0.5 & 1.0 & 2.0 & 4.0\\
	\midrule
	PointPillars~\cite{lang2019pointpillars} & 49.7 & 59.2 & 63.7 & 66.9 \\
	PointPillars+~\cite{vora2020pointpainting} & 56.5 & 63.7 & 67.5 & 70.0 \\
	3DSSD~\cite{yang20203dssd} & 65.8 & 69.0 & 71.7 & 74.1 \\
	PointPainting~\cite{vora2020pointpainting} & 64.3 & 72.5 & 76.7 & 79.3 \\
	MegVII~\cite{zhu2019classbalanced} & 78.8 & 80.3 & 81.9 & 83.6 \\
	CenterPoint~\cite{yin2020center} & 82.2 & 84.1 & 85.5 & 86.9 \\
	\midrule
    Our Backbone & 72.1 &	72.7 & 73.1 & 73.9 \\
    \bottomrule
    \end{tabular}
    \caption{\textbf{[\nuscenes{}] Detection results}. Comparison of our backbone network (without map) against state-of-the-art detectors.}
    \cutcaptiondown
    \cutcaptiondown
    \label{table:nuscenes_detection}
\end{table}

\nuscenes includes 3D point clouds from a 32-beam LiDAR sensor and HD annotated maps.
The sparse point clouds of this dataset make perception more challenging. 
The dataset consists of scenes in Boston and Singapore, and importantly, includes over 200,000 annotated pedestrians. 
However, we find that the tracks have short duration, and thus we limit the task to 4 second forecasting.
We evaluate results on the held out validation set as the test set has no prediction labels and their benchmark is different from ours, \ie, prediction given perfect perception vs. our joint perception and prediction. We recall that the problem we are trying to tackle is precisely how to be safe when having a real perception system that is not perfect.

In Table~\ref{table:results_nuscenes} and Fig.~\ref{fig:scene_nuscenes}, we present quantitative comparisons to the baselines on this dataset.
The results show that our method achieves better performance than previous state-of-the-art motion forecasting methods on Scene Occupancy while being on par in Motion Forecasting. Please note that the small differences in Motion Forecasting are not significant, and there is no model that outperforms the rest.
In particular, we observe great gains in Scene Occupancy Precision-Recall (PR) metrics (Avg. and Final mAP). These are the most important metrics as they measure the ranking of the scene occupancy scores. Our method also outperforms all baselines in the Maximum Calibration Error (MCE), and is only surpassed by SocialSTGCNN~\cite{mohamed2020social} in the Average Calibration Error (ACE). We consider MCE more relevant to safety-critical applications such as driving, since it captures the worst case, i.e. largest mismatch between confidence and accuracy.
Note that the absolute numbers in \ourdataset and \nuscenes are quite different because the scene is $200\times200$ in \nuscenes and $100\times100$ in \ourdataset, and the prediction horizon is 4s in \nuscenes and 7s in \ourdataset.

Table \ref{table:nuscenes_detection} shows a comparison of our perception network against state-of-the-art detectors.
The goal of these results is to showcase that our perception network architecture employed both in our model as well as the baselines is competitive with state-of-the-art detectors in the \nuscenes detection challenge. This is important because the object detector needs to work fairly well for the actor motion evaluation to be reasonable, as this is evaluated in true positive detections. However, we emphasize that this is not our contribution, but rather a sanity check.
We also note that to show a fair comparison to the baselines we remove the map stream of our backbone network in this experiment as the detection benchmark does not allow the use of maps.

\begin{figure*}[h]
    \centering
    \begin{tabular} {@{}c@{\hspace{.5em}}c@{\hspace{.5em}}c@{\hspace{.5em}}c}
        \rotatebox[origin=c]{90}{\textbf{STGAT}} & \raisebox{-0.5\height}{\includegraphics[width=0.32\textwidth, trim={0.1cm, 5.0cm, 0.1cm, 3.0cm}, clip]{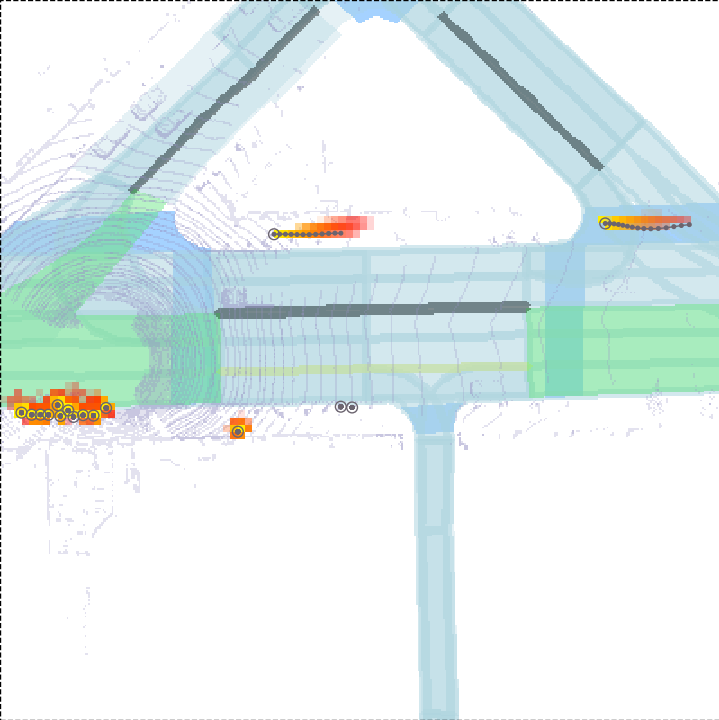}} &
        \raisebox{-0.5\height}{\includegraphics[width=0.32\textwidth, trim={0.1cm, 5.0cm, 0.1cm, 3.0cm}, clip]{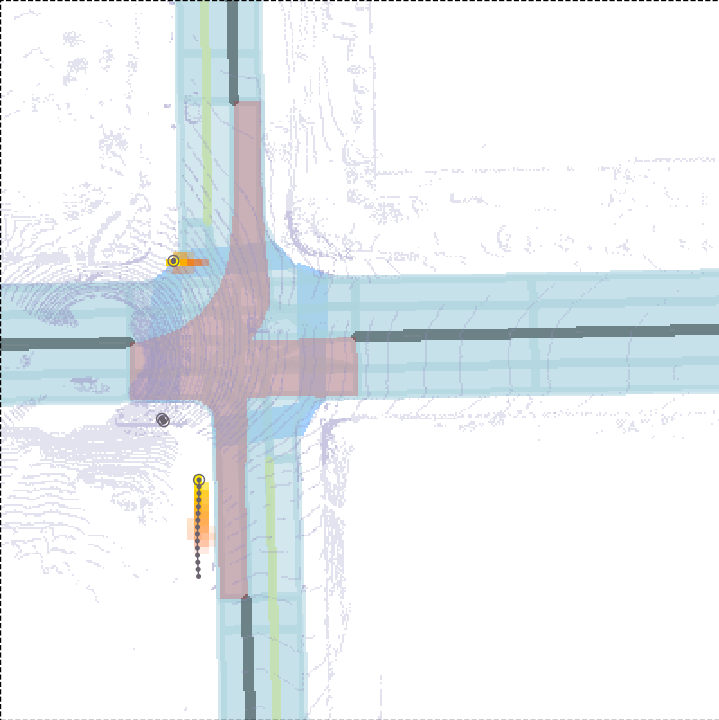}} & 
        \raisebox{-0.5\height}{\includegraphics[width=0.32\textwidth, trim={0.1cm, 5.0cm, 0.1cm, 3.0cm}, clip]{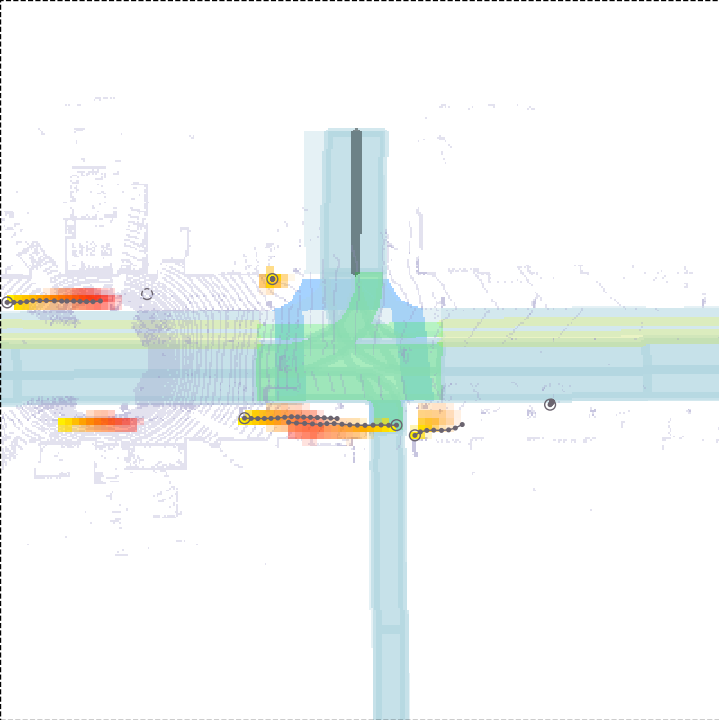}} \vspace{.5em} \\
        \midrule
        \rotatebox[origin=c]{90}{\textbf{SocialSTGCNN}} & \raisebox{-0.5\height}{\includegraphics[width=0.32\textwidth, trim={0.1cm, 5.0cm, 0.1cm, 3.0cm}, clip]{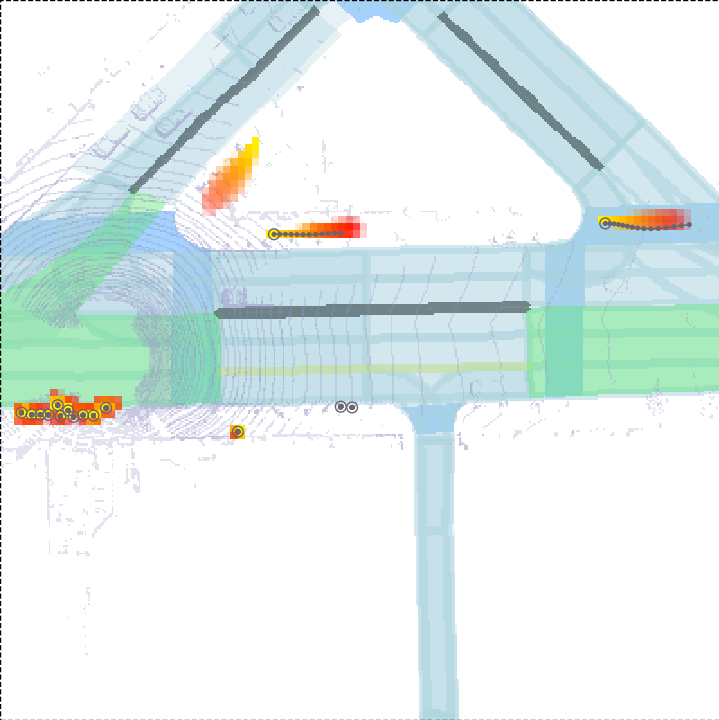}} &
        \raisebox{-0.5\height}{\includegraphics[width=0.32\textwidth, trim={0.1cm, 5.0cm, 0.1cm, 3.0cm}, clip]{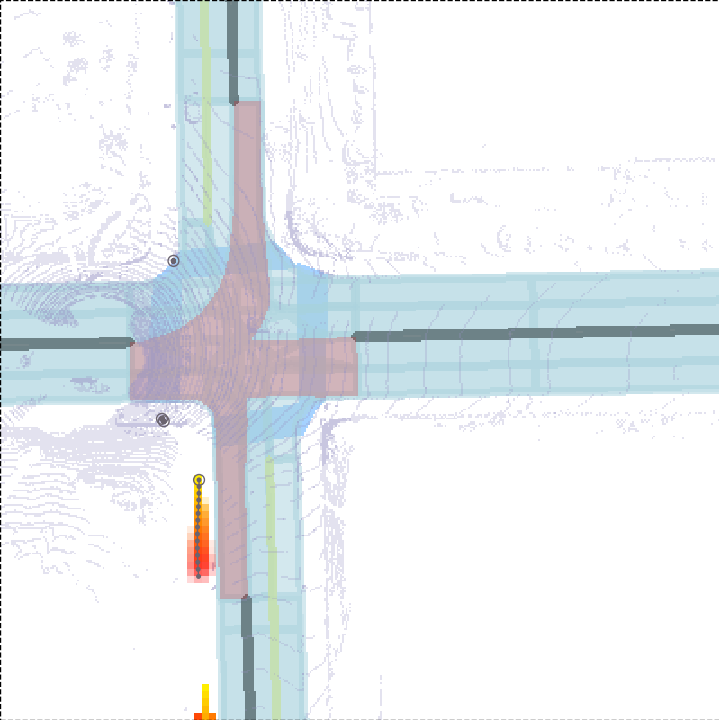}} &
        \raisebox{-0.5\height}{\includegraphics[width=0.32\textwidth, trim={0.1cm, 5.0cm, 0.1cm, 3.0cm}, clip]{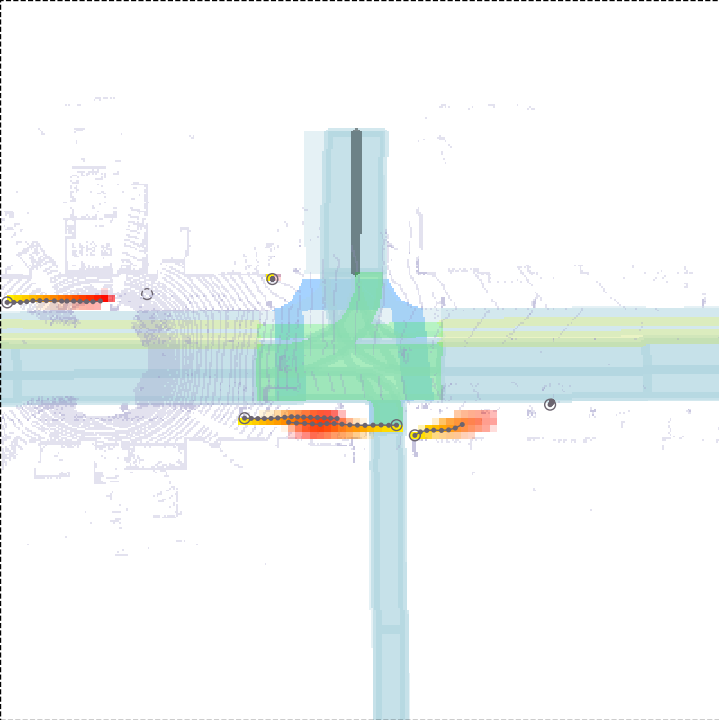}} \vspace{.5em} \\
        \midrule
        \rotatebox[origin=c]{90}{\textbf{SpAGNN}} & \raisebox{-0.5\height}{\includegraphics[width=0.32\textwidth, trim={0.1cm, 5.0cm, 0.1cm, 3.0cm}, clip]{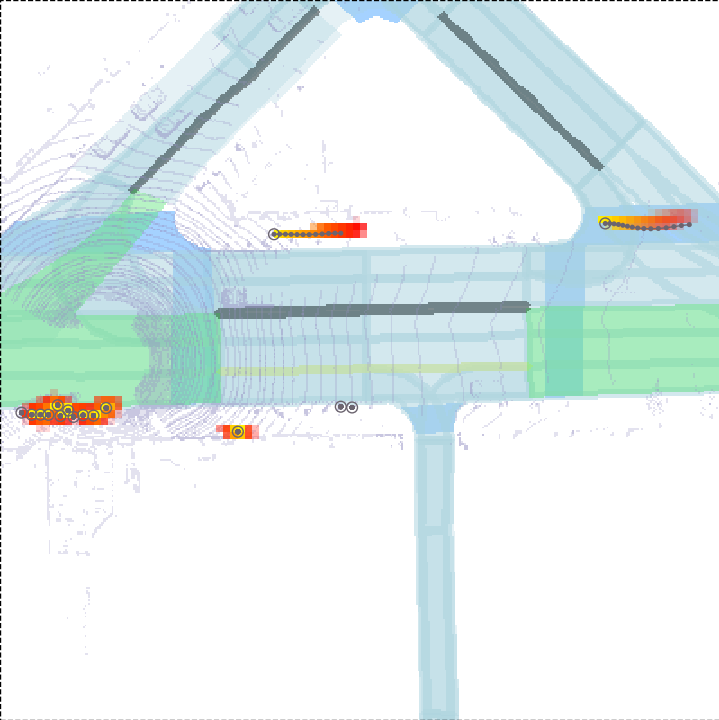}} &
        \raisebox{-0.5\height}{\includegraphics[width=0.32\textwidth, trim={0.1cm, 5.0cm, 0.1cm, 3.0cm}, clip]{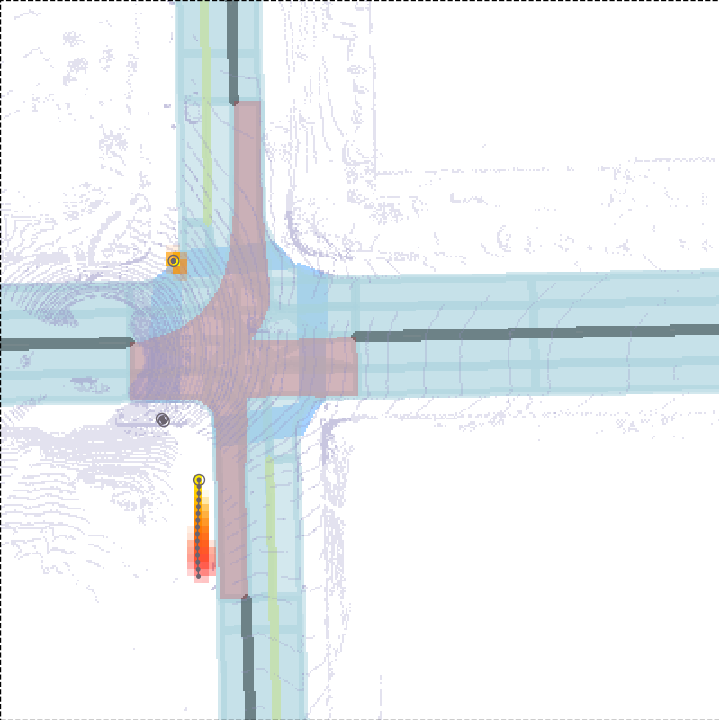}} & 
        \raisebox{-0.5\height}{\includegraphics[width=0.32\textwidth, trim={0.1cm, 5.0cm, 0.1cm, 3.0cm}, clip]{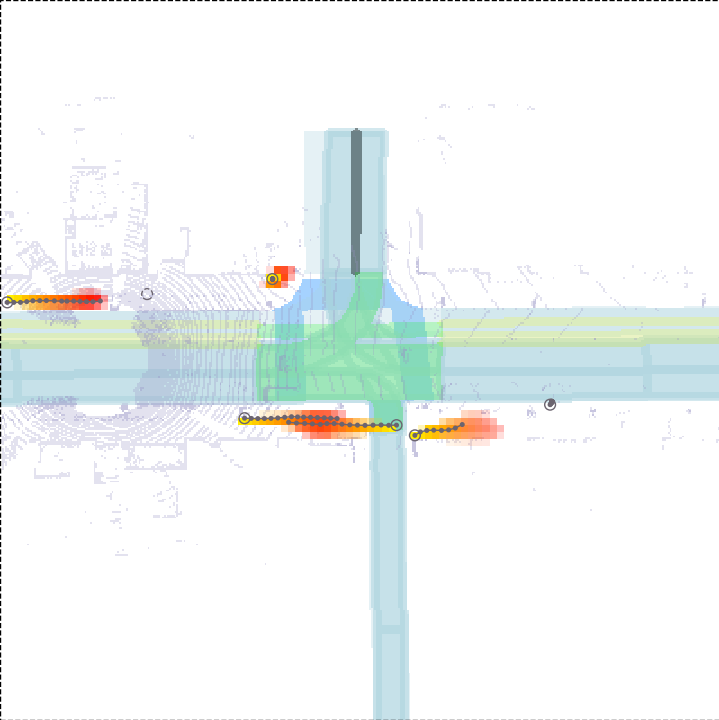}} \vspace{.5em} \\
        \midrule
        \rotatebox[origin=c]{90}{\textbf{MultiPath}} & \raisebox{-0.5\height}{\includegraphics[width=0.32\textwidth, trim={0.1cm, 5.0cm, 0.1cm, 3.0cm}, clip]{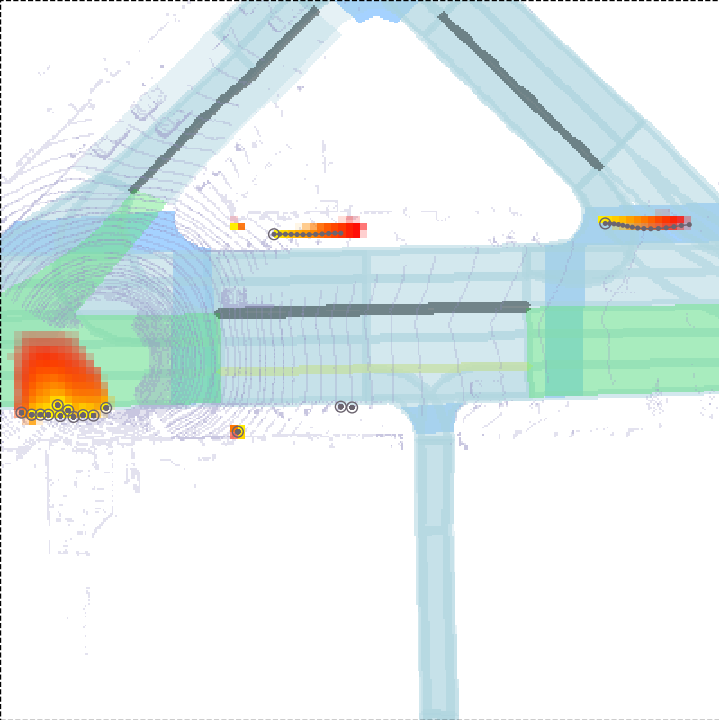}} &
        \raisebox{-0.5\height}{\includegraphics[width=0.32\textwidth, trim={0.1cm, 5.0cm, 0.1cm, 3.0cm}, clip]{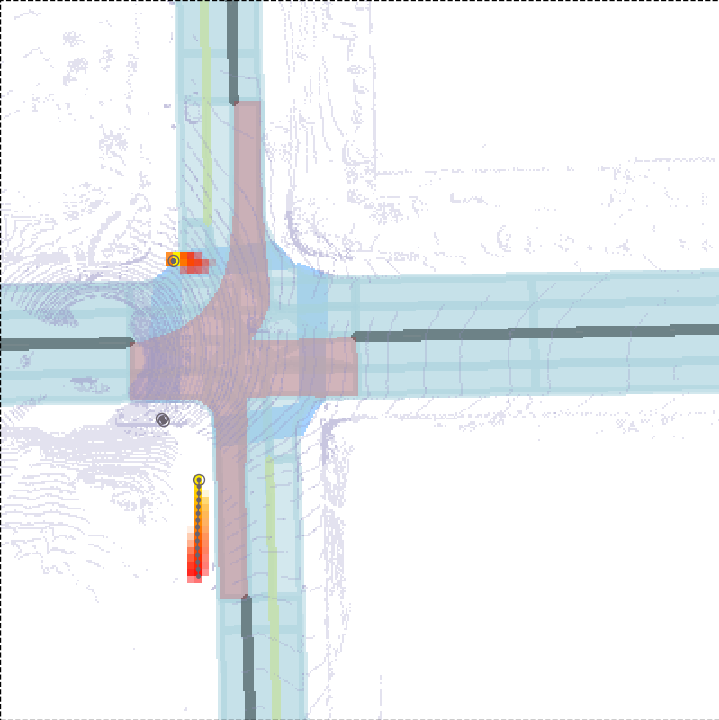}} & 
        \raisebox{-0.5\height}{\includegraphics[width=0.32\textwidth, trim={0.1cm, 5.0cm, 0.1cm, 3.0cm}, clip]{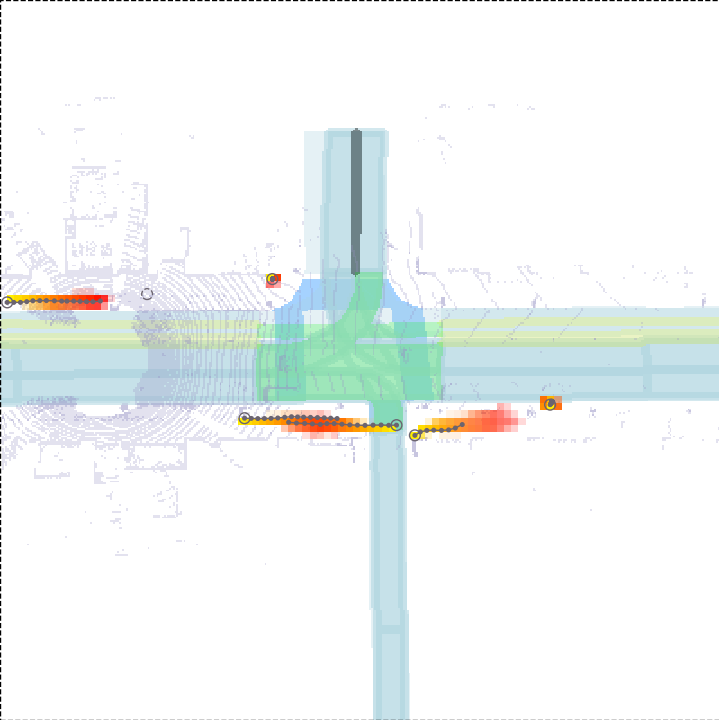}} \vspace{.5em} \\
        \midrule
        \rotatebox[origin=c]{90}{\textbf{DRF-Net}} & \raisebox{-0.5\height}{\includegraphics[width=0.32\textwidth, trim={0.1cm, 5.0cm, 0.1cm, 3.0cm}, clip]{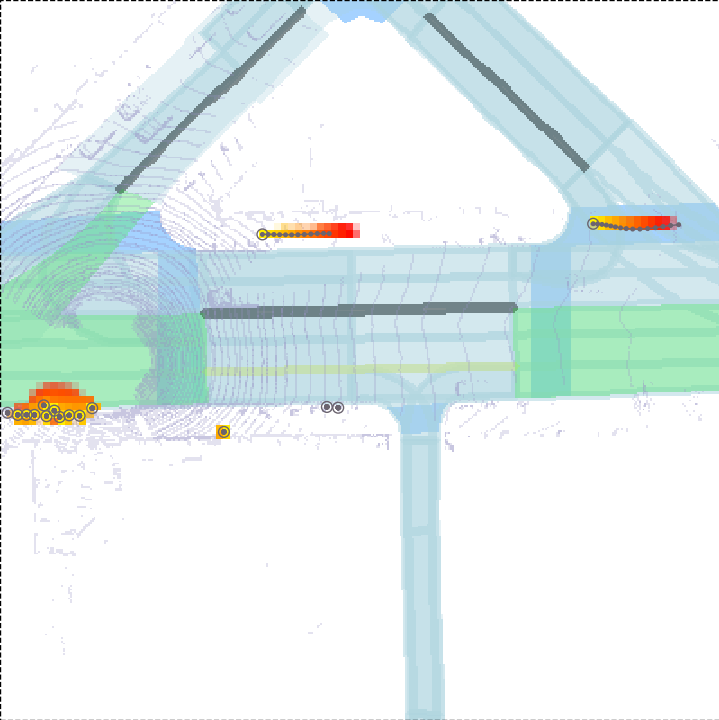}} &
        \raisebox{-0.5\height}{\includegraphics[width=0.32\textwidth, trim={0.1cm, 5.0cm, 0.1cm, 3.0cm}, clip]{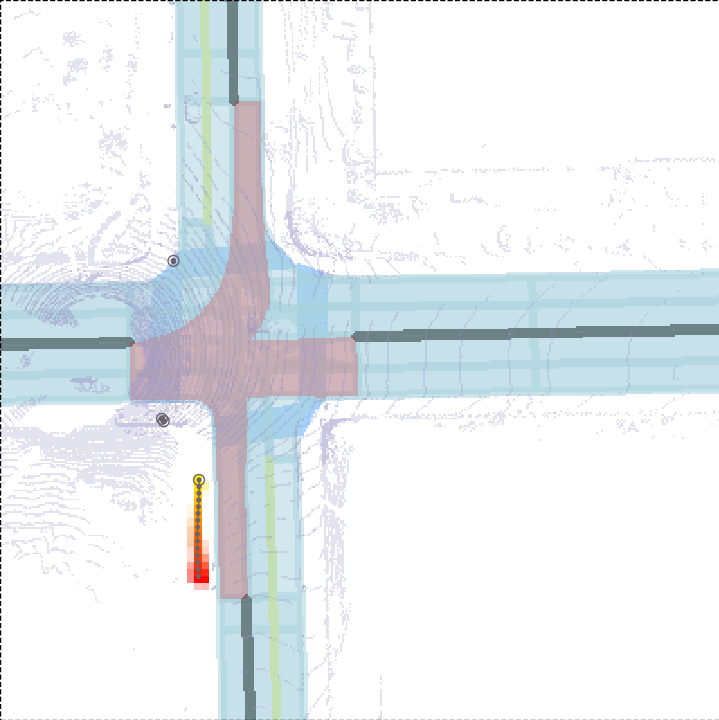}} & 
        \raisebox{-0.5\height}{\includegraphics[width=0.32\textwidth, trim={0.1cm, 5.0cm, 0.1cm, 3.0cm}, clip]{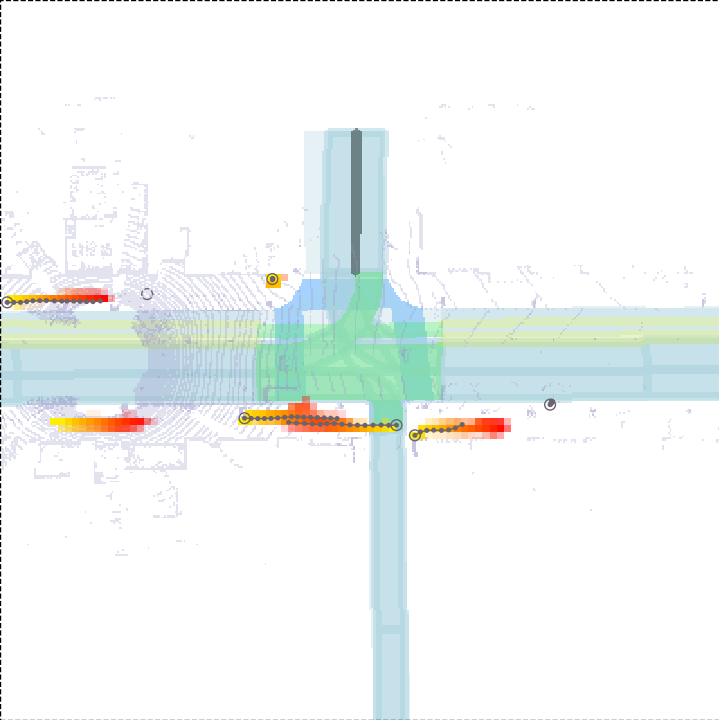}} \vspace{.5em} \\
        \midrule
        \rotatebox[origin=c]{90}{\textbf{SA-GNN (Ours)}} & \raisebox{-0.5\height}{\includegraphics[width=0.32\textwidth, trim={0.1cm, 5.0cm, 0.1cm, 3.0cm}, clip]{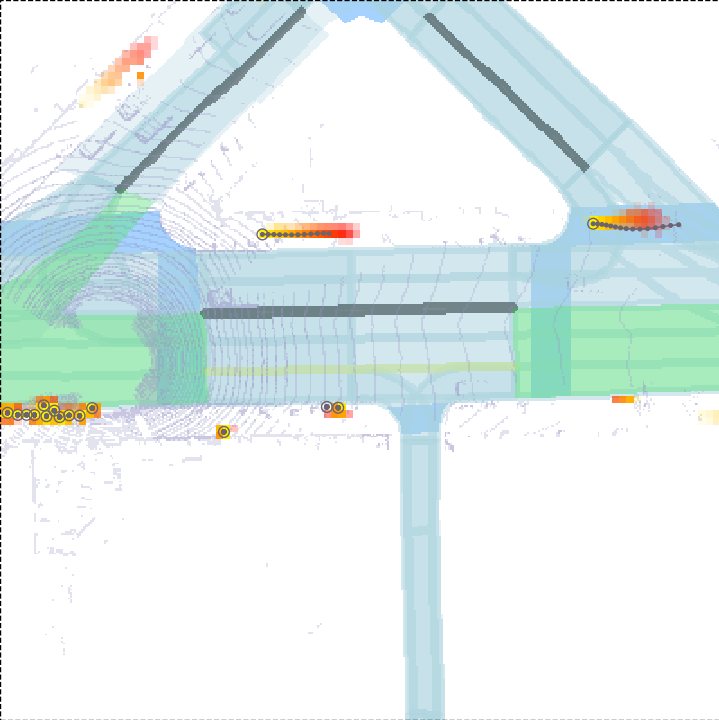}} &
        \raisebox{-0.5\height}{\includegraphics[width=0.32\textwidth, trim={0.1cm, 5.0cm, 0.1cm, 3.0cm}, clip]{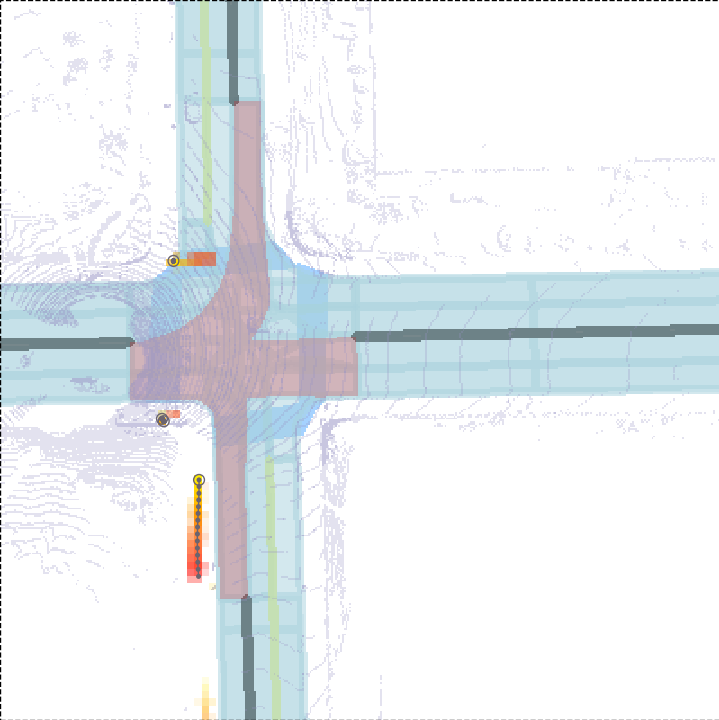}} & 
        \raisebox{-0.5\height}{\includegraphics[width=0.32\textwidth, trim={0.1cm, 5.0cm, 0.1cm, 3.0cm}, clip]{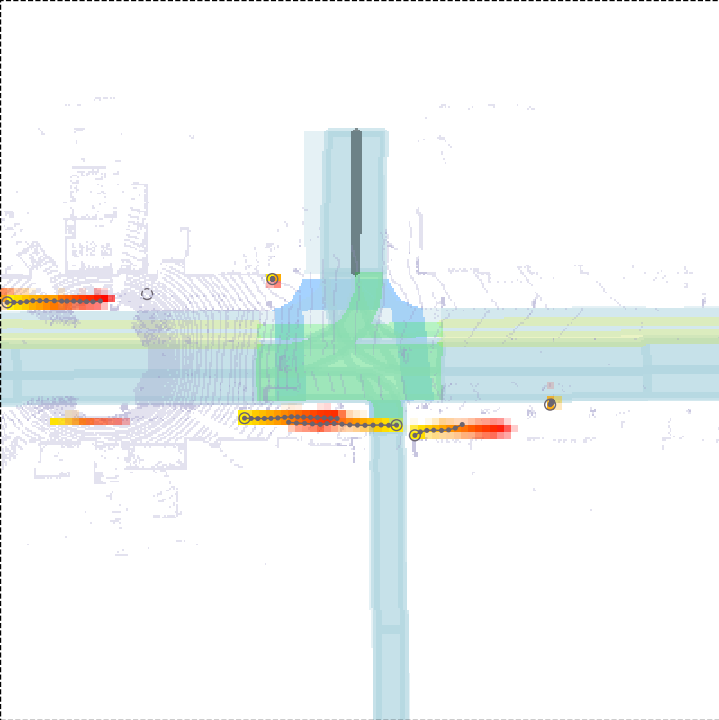}} \vspace{.5em} \\
    \end{tabular}
    \cutcaptionup
    \caption{\textbf{[\ourdataset{}] Qualitative scene occupancy comparison} with baselines. We highlight the high recall of \ourmodelshort's scene occupancy prediction, which does not miss any ground-truth pedestrian that is partially visible (solid gray circles). In contrast, the baselines miss pedestrians near the curb due to detection thresholding that could be critical for the SDV's decision making.
    }
    \cutcaptiondown
    \label{fig:qualitative_supp_so}
\end{figure*}

\end{document}